\pdfoutput=1
\documentclass[journal]{IEEEtran}
\usepackage{amsmath,amsfonts}
\usepackage{algorithmic}
\usepackage{algorithm}
\usepackage{array}
\usepackage[caption=false,font=normalsize,labelfont=sf,textfont=sf]{subfig}
\usepackage{textcomp}
\usepackage{stfloats}
\usepackage{url}
\usepackage{verbatim}
\usepackage{graphicx}
\usepackage{cite}
\usepackage{color}
\usepackage{makecell}
\usepackage{multirow}
\usepackage{float}  
\usepackage{subfig}
\usepackage{amsthm}
\usepackage{threeparttable}
\newcommand{\mz}[1]{\textcolor{black}{#1}}
\newcommand{\zh}[1]{\textcolor{black}{#1}}
\newtheoremstyle{MyThmStyle}
{}
{}
{\itshape}
{}
{\bfseries}
{}
{ }
{\thmname{#1\thmnumber{ #2\hspace{0.5em}}}\thmnote{(#3)}}
\theoremstyle{MyThmStyle}
\newtheorem{theo}{Proposition}
\newtheorem{mark1}{Remark}


\usepackage{tikz,xcolor}
\usepackage[implicit=false]{hyperref}

\hypersetup{hidelinks,
	colorlinks=true,
	allcolors=black,
	pdfstartview=Fit,
	breaklinks=true}
	
\definecolor{lime}{HTML}{A6CE39}
\DeclareRobustCommand{\orcidicon}{
\begin{tikzpicture}
\draw[lime, fill=lime] (0,0)
circle[radius=0.16]
node[white]{{\fontfamily{qag}\selectfont \tiny \.{I}D}};
\end{tikzpicture}
\hspace{-2mm}
}
\foreach \x in {A, ..., Z}{%
\expandafter\xdef\csname orcid\x\endcsname{\noexpand\href{https://orcid.org/\csname orcidauthor\x\endcsname}{\noexpand\orcidicon}}
}

\usepackage{algorithmic}
\usepackage{algorithm}

\begin{document}

\title{Gradually Vanishing Gap in Prototypical Network \\for Unsupervised Domain Adaptation}

\author{Shanshan Wang, 
        Hao Zhou, 
        Xun Yang,
        Zhenwei He,
        Mengzhu Wang,
        Xingyi Zhang,
        Meng Wang  
        
\thanks{Shanshan Wang and Hao Zhou are with the Information Materials and Intelligent Sensing Laboratory of Anhui Province, Institutes of Physical Science and Information Technology, Anhui University, Hefei 230601, China.   (e-mail: wang.shanshan@ahu.edu.cn,zhouhaokey852@gmail.com).}
\thanks{Xun Yang is with the Department of Electronic Engineering and Information Science, School of Information Science and Technology, University of Science and Technology of China, Hefei 230026, China (e-mail: xyang21@ustc.edu.cn)}
\thanks{Zhenwei He is with the College of Computer Science and Engineering, Chongqing University of Technology, Chongqing 400054, China (e-mail: hzw@cqut.edu.cn)}
\thanks{Meng Wang is with the School of Computer Science and Information Engineering, Hefei University of Technology, Hefei 230009, China (e-mail:wangmeng@hfut.edu.cn)}
\thanks{Mengzhu Wang is with the School of Artificial Intelligence, Hebei University of Technology, Tianjin, P.R. China (e-mail: dreamkily@gmail.com)}
\thanks{Xingyi Zhang is with the Key Laboratory of Intelligent Computing and Signal Processing, Ministry of Education, and the School of Computer Science and Technology, Anhui University, Hefei 230601, China (e-mail: xyzhanghust@gmail.com).}
}


 
\maketitle

\begin{abstract}
Unsupervised domain adaptation~(UDA) is a critical problem for transfer learning, which aims to transfer the semantic information from labeled source domain to unlabeled target domain. Recent advancements in UDA models have demonstrated significant generalization capabilities on the target domain. However, the generalization boundary of UDA models remains unclear. When the domain discrepancy is too large, the model can not preserve the distribution structure, leading to distribution collapse during the alignment. To address this challenge, we propose an efficient UDA framework named \textbf{G}radually \textbf{V}anishing \textbf{G}ap in \textbf{P}rototypical \textbf{N}etwork~(GVG-PN), which achieves transfer learning from both global and local perspectives. From the global alignment standpoint, our model generates a domain-biased intermediate domain that helps preserve the distribution structures. By entangling cross-domain features, our model progressively reduces the risk of distribution collapse. However, only relying on global alignment is insufficient to preserve the distribution structure. To further enhance the inner relationships of features, we introduce the local perspective. We utilize the graph convolutional network (GCN) as an intuitive method to explore the internal relationships between features, ensuring the preservation of manifold structures and generating domain-biased prototypes.  Additionally, we consider the discriminability of the inner relationships between features. We propose a pro-contrastive loss to enhance the discriminability at the prototype level by separating hard negative pairs. By incorporating both GCN and the pro-contrastive loss, our model fully explores fine-grained semantic relationships.  Experiments on several UDA benchmarks validated that the proposed GVG-PN can clearly outperform the SOTA models.
\end{abstract}


\section{Introduction}
\IEEEPARstart{R}{ecently}, with the development of deep convolutional neural networks~(CNNs)~\cite{he2016deep}, many computer vision models have achieved outstanding performance based on abundant labeled data. However, the performance of these models is often affected by distribution discrepancies between different datasets. \textit{e.g.}, sketches often lack detailed color information, whereas real-world photos exhibit rich colors.
Due to the domain bias, the networks trained on sketches do not always perform well on real-world photos~\cite{pan2009survey}. 
In recent years, unsupervised domain adaptation~(UDA) has emerged as the mainstream method to address the domain gap issue and it aims to transfer the knowledge learned from the labeled source domain to the unlabeled target domain. 
\zh{In our study, the downstream task revolves around image classification. UDA leverages the labeled source domain samples to train a classification network with robust generalization.
This allows the network to exhibit optimal classification predictions even when confronted with target domain samples lacking labels.
}

\begin{figure}[t]
    \centering     \includegraphics[width=0.9\linewidth]{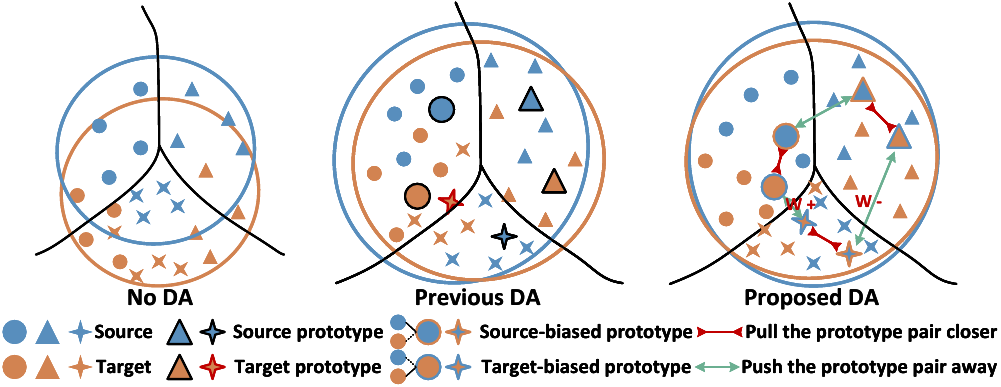}   
    \caption{
    \zh{Motivation for the proposed approach. 
        Previous DA methods that directly align two domains can not prevent the misclassification of target samples. In some cases, prototypes of certain categories may stay in incorrect category spaces.
        To overcome this issue, our proposed approach aims to generate two intermediate domains to achieve progressive alignment.
        By exploring both global and local distributions, we ensure fine-grained semantic relationships during the generation of intermediate domains.
        Prototypes are utilized to describe the semantic structure of these intermediate domains.
        The parameter 'w' represents the extent to which prototypes push apart, thereby enhancing the discriminative ability of hard alignment on categories.
        Consequently, our model is capable of aligning different distributions while maintaining the integrity of the distribution structure.}
    }
    \label{domain}
\end{figure}

Most UDA methods~\cite{tzeng2014deep,wang2021self,co-hhda,wangmm,medm,wangtip,wangtnnls,tflgm,tnnls1,Zheng2023detection,Li2023re-id} aim to reduce the distribution discrepancy between the features of two domains by mapping them into a common feature space. Generally, these methods can be classified into two categories: statistical-based methods and adversarial-based methods. Statistical-based methods typically employ distance metrics such as MMD~\cite{tzeng2014deep,long2015learning,long2016unsupervised,long2017deep}, KL-Divergence~\cite{zhuang2015supervised} and Wasserstein distance~\cite{shen2018wasserstein} to measure and minimize the statistical discrepancy between the two domains. On the other hand, adversarial-based methods~\cite{ma2019gcan,ganin2016domain,saito2018maximum,xie2018learning,long2018conditional,sharma2021instance} focus on learning domain invariant representations through adversarial learning, which involves a minimax game between the feature extractor and the domain discriminator. 
However, in cases where the domain gap is excessively large, aligning the features becomes challenging due to the collapse of the distribution structure. This is because some previous methods overlook the semantic relationships between features, leading to suboptimal models. 
\zh{
Additionally, in Figure.~\ref{domain}, conventional DA methods may face challenges when adapting to difficult categories, where the features from the two categories are similar. This could result in the prototype of that target category (the mean of all features of the category) being positioned in an incorrect category space, leading to a significant number of mispredictions for the majority of samples in that category.}

\begin{figure}[t]
    \centering     \includegraphics[width=0.99\linewidth] {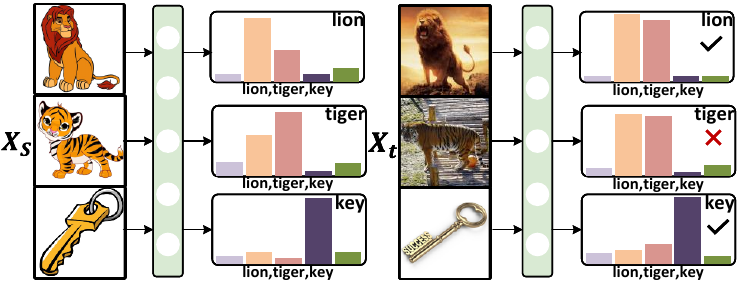}    
    \caption{Motivation of the pro-contrastive learning. Although the transferability in DA could alleviate the domain discrepancy problem, it may not effectively address misclassification issues with hard samples. We aim to assign greater  weight to these challenging hard class pairs. By doing so, the discriminability of the hard negative pairs is enhanced, leading to improved separation of features between classes such as 'lion' and 'tiger'.
    }
    \label{domain2}
\end{figure}

To address the aforementioned problem, in this paper, we propose a novel method named Gradually Vanishing Gap in Prototypical Networks~(GVG-PN), which focuses on aligning feature distributions from both global and local perspectives. 
\zh{Different from the majority of previous methods~\cite{kang2019contrastive,xie2018learning} that directly perform global and local alignment,}
instead of directly aligning the original domains, our approach aims to achieve domain adaptation~(DA) on two generated intermediate domains at a global level. Specifically, 
considering that directly adapting the domain features may yield suboptimal results due to significant domain discrepancies, we aim to overcome the limitations of the uncertain DA boundary by constructing two intermediate domains. With the progressive domain alignments, our model can preserve the original semantic relation during the alignment, thereby the risk of distribution collapse is reduced.

However, considering the possibility of a significant domain shift, directly applying the global alignment is not enough, it is important to preserve the fine-grained manifold structures and achieve the local alignment process. To address this, it is necessary to incorporate the preservation of semantic structures within the model. 
Inspired by~\cite{luo2020progressive,roy2021curriculum}, 
we adopt the graph convolutional network~(GCN) as it is a great method for keeping the inner feature structure.
Specifically, leveraging the benefits of the GCN, features are aggregated considering both intra-domain and inter-domain relationships, enabling the generation of source-biased and target-biased prototypes respectively. With the entanglement of cross-domain features, our model can alleviate the impact of large domain discrepancies and progressively achieve the DA alignment as presented in Figure.~\ref{domain}.

\zh{It is worth noting that, in contrast to previous methods~\cite{na2021fixbi, inter} for generating the intermediate domain, our approach constructs the intermediate domain based on feature relationships.
During the training process, we employ a single layer GCN to learn the semantic similarity among all samples within a batch. Subsequently, we aggregate sample features based on this similarity to generate the feature representation of the intermediate domain.
The domain-biased prototype aggregates sample features from two domains and encapsulates the semantic structure of the intermediate domain.
In detail, a domain-biased prototype for a specific class encompasses a significant number of sample features from that class within the domain, as well as incorporating sample features from the same class in another domain.
}

In the process of domain adaptation, we generated intermediate domain that reduced differences in the global domain, while utilizing the GCN to preserve local manifold structures. However, relying solely on transferability is not enough and the class-wise discriminability is also crucial. 
Intuitively,  features belonging to the same class should be close together, while different classes should be separated as much as possible.
To address this, we employ contrastive learning~\cite{wu2018unsupervised,he2020momentum,chen2020simple} to obtain discriminative features.
Traditional contrastive learning treats all sample pairs equally, which is not suitable when there are class-wise similarity imbalances in semantic relations as shown in Figure.~\ref{domain2}. For instance, compared with the 'lion’, the ‘tiger’ exhibits higher similarity than the 'key' obviously. Consequently, the model has a greater chance of misclassifying 'tiger' as 'lion' rather than 'key'. In such cases, prototypes with similar appearances become 'hard' pairs to be separated, while prototypes with large discrepancies are considered 'easy' pairs.
\zh{As illustrated in Figure~\ref{domain},}
in response to this issue, we aim to dedicate more weights to these harder prototype pairs, enhancing the discriminability of the hard negative pairs. 
In this paper, based on the obtained domain-biased prototypes, we construct a pro-contrastive loss to train the prototypes to be far away from each other. Specifically, we introduce an anchor-based weighted mechanism in the loss to make the model self-adaptively assign more weight to the harder prototype pairs, ensuring inner discriminability during the domain alignment. 

\zh{
In conclusion, this paper proposes a progressive DA model that gradually aligns the original domain by aligning the intermediate domain. 
Our approach utilizes GCN to learn fine-grained semantic relationships among samples, then the domain-biased prototypes are generated by following the clustered features.  
}
Firstly, the prototypes are learned not only from other related classes but also from other domains. Secondly, due to its natural property, features can preserve the original manifold structures. Additionally, considering that relying only on transferability may not fully explore the inner relations between different classes, we introduce a novel constructive learning paradigm called pro-contrastive loss to explore fine-grained discriminative features. 
To summarize, the contributions of our paper can be summarized as follows:
\begin{itemize}
\item {To address the large domain gap problem, our approach introduces a framework that achieves DA from both global and local perspectives. On one hand, our method focuses on reducing the domain gap globally, while on the other hand, we preserve the local manifold structures to avoid distribution collapse.
}

\item {Instead of aligning the two original domains directly, we adopt a strategy to progressively align two generated intermediate domains. This approach allows us to leverage the GCN to not only preserve the manifold structures but also aggregate domain features into two domain-biased prototypes.
}
\item {In order to fully explore the fine-grained semantic relations, as well as transferability, we introduce a pro-contrastive loss to enhance class-level discriminability. Specifically, this loss focuses on the hard negative pair, constraining the prototypes of hard negative pairs to be farther apart. By doing so, our model can overcome the limitations of domain gaps and achieve both transferability and discriminability.}
\end{itemize}

The remaining parts of this paper are organized as follows.
Section II briefly describes the relevant work in our study.
Section III introduces the progressive alignment framework proposed and focuses on domain-biased prototype modeling. 
Section IV presents the experimental results and compares them with state-of-the-art UDA methods.
In Section V, a detailed experimental analysis is conducted to demonstrate the effectiveness of the proposed method GVG-PN.
Finally, Section VI concludes the paper. 

\section{RELATED WORK}
In this section, we will review the related work in UDA, as well as other involved research within our framework,~\textit{i.e.}, graph neural networks~(GNN) and contrastive learning.

\subsection{Unsupervised Domain Adaptation}
In recent years, UDA has emerged as a popular research direction in computer vision, with intensive exploration conducted by various researchers~\cite{wang2021self,co-hhda,wangmm,medm}.
Generally, these methods can be classified into two categories: statistical-based and adversarial-based approaches.
Several methods~\cite{ben2010theory,wang2023reducing} utilize the Maximum Mean Discrepancy~(MMD) ~\cite{tzeng2014deep,long2015learning,long2016unsupervised,long2017deep} as a distance metric to align the feature distributions.
DAN~\cite{long2015learning} employs the multiple kernel variant of MMD~(MK-MMD) to adapt the task-specific final layers.
JAN~\cite{long2017deep} aligns the marginal and joint distribution discrepancies between domains, while RTN~\cite{long2016unsupervised} incorporates residual functions into the model to mitigate domain shift. 
Moreover, statistical metrics like CORAL~\cite{sun2016deep}, KL-Divergence~\cite{zhuang2015supervised}, and Wasserstein distance~\cite{shen2018wasserstein} are extensively utilized in UDA. In addition to transferability, discriminability plays a crucial role in domain adaptation tasks.
To exploit categorical information, methods like CAN~\cite{kang2019contrastive} construct category-aware UDA networks at the class level. When dealing with a significant domain gap,
FixBi~\cite{na2021fixbi} leverages data augmentation to dynamically generate multiple intermediate domains, aiming to mitigate negative transfer problems. While our method shares similarities with FixBi, we differentiate ourselves by leveraging the original data instead of generating new data.
Another mainstream approach in UDA is adversarial methods~\cite{ma2019gcan,ganin2016domain,saito2018maximum,xie2018learning,long2018conditional,sharma2021instance,wang2023bp,wang2022informative}, inspired by generative adversarial networks~(GAN)~\cite{goodfellow2020generative}.
DANN~\cite{ganin2016domain} is a classical adversarial-based DA method that achieves adversarial learning by incorporating a gradient reversal layer. Similarly, MSTN~\cite{xie2018learning}  aligns category prototypes between two domains, similar to our approach. However, we differ in our strategy by aligning the category prototypes of two intermediate domains instead.

\subsection{Graph Neural Networks}
With the aid of Graph Neural Networks~ (GNN)\cite{scarselli2008graph}, it becomes possible to learn from unstructured data by constructing graph structures. Consequently, numerous research studies have emerged in this field~\cite{kipf2016gcn,hamilton2017inductive,velickovic2017graph}.
GraphSAGE~\cite{hamilton2017inductive} is capable of generating embeddings for unknown nodes through sampled node embeddings.
In the context of message passing, the graph attention network~\cite{velickovic2017graph} assigns varying weights to neighboring nodes, thereby considering the importance of each node.
In UDA, GNN has been widely leveraged to tackle the problem of domain shift.
PGL~\cite{luo2020progressive} introduces a progressive GCN framework to address domain shift in OSDA.
To adapt to multi-target domains, D-CGCT~\cite{roy2021curriculum} utilizes GNN for feature aggregation while incorporating co-teaching and curriculum learning for holistic network training.
GCAN~\cite{ma2019gcan} emphasizes the significance of inter-domain data structures, constraining graph networks to obtain structure scores for domain alignment using triplet loss~\cite{hermans2017defense}.
ILA-DA~\cite{sharma2021instance} constructs an affinity matrix between samples to regulate intra-class and inter-class relationships, serving as an inspiration for our adoption of GCN in our method. However, unlike these approaches, we employ GCN not only to preserve local structures but also to generate global domain-biased prototypes considering both intra-sample and inter-sample relationships.
\subsection{Contrastive Learning}
Recently, contrastive learning~\cite{wu2018unsupervised,he2020momentum,chen2020simple,oord2018representation} has received significant attention and has made remarkable progress. The objective of contrastive learning is to bring positive pairs closer together while pushing negative pairs further apart. 
Initially, contrastive learning employed individual discriminative tasks~\cite{wu2018unsupervised} as proxy tasks for model pre-training. 
SimCLR~\cite{chen2020simple} popularized the approach by using data augmentation samples from the same image as positive pairs and samples from different images as negative pairs. This became the mainstream method in contrastive learning.
Integrating contrastive learning into the UDA framework has proven to be effective, as downstream tasks such as classification and semantic segmentation require discriminative semantic information for optimal performance. 
CDA~\cite{thota2021contrastive} independently utilized contrastive loss in both domains and eliminated false negative samples to enhance model performance.
HCL~\cite{huang2021model}, operating in the source-free adaptation setting, introduced a historical contrastive learning framework to compensate for the absence of source data. 
CaCo~\cite{huang2022category}  incorporated a semantic prior through instance contrastive learning, constructing a semantic-aware dictionary to facilitate key pair querying.
In our approach, we construct a variant of contrastive loss to enhance the discriminability of challenging hard pairs, thereby improving performance in the final task.

\begin{figure*}[t]
    \centering  
    \includegraphics[width=0.9\linewidth] {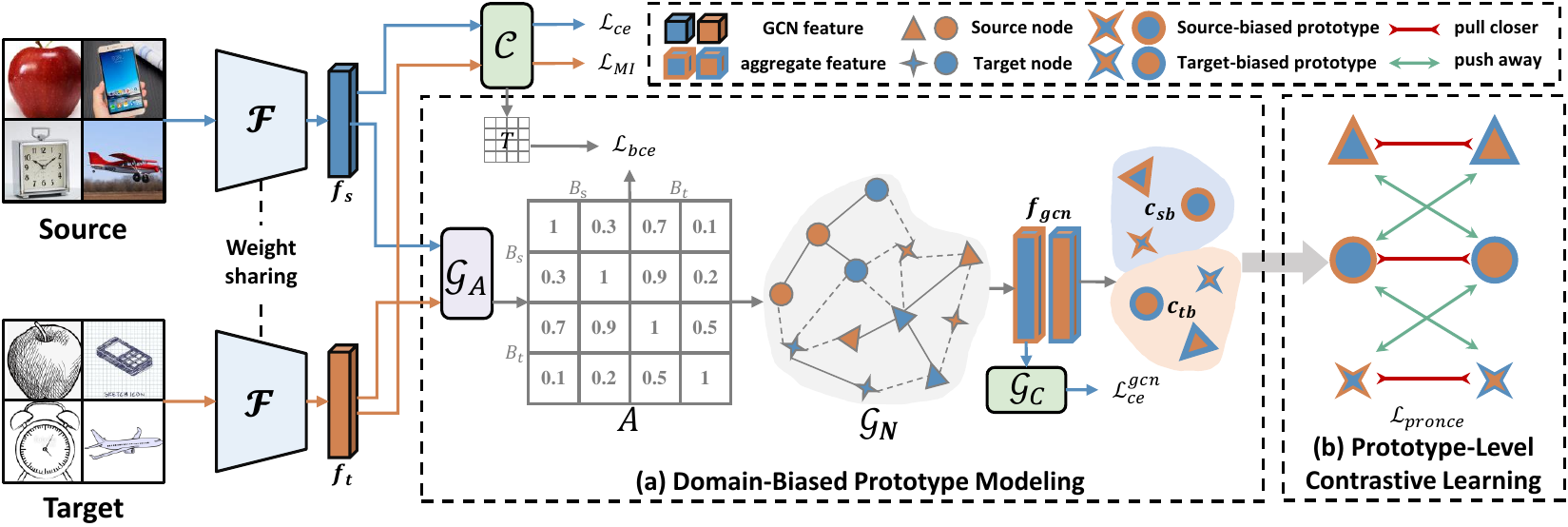} 
    \caption{An overview of our GVG-PN method is presented as follows.
    \zh{$\mathcal{F}$ signifies the feature extractor, $\mathcal{C}$ and $\mathcal{G}_C$ represent the classifier components, $\mathcal{G}_A$ denotes the affinity matrix generation layer, $\mathcal{G}_N$ denotes the node update layer, and $T$ corresponds to the ground-truth matrix.}
    ~(a)  In the feature aggregation phase, the ground-truth label guides $\mathcal{G}_{A}$ to generate the affinity matrix $A$. Subsequently, the node features are fed into $\mathcal{G}_{N}$ to obtain the aggregated features ${f}_{gcn}$. To generate domain-biased prototypes, we compute prototypes for each category based on the aggregated features. During the prototype generation process, both intra-class and inter-class relationships are taken into consideration.
    ~(b) We utilize the prototypes to explore the discriminability of classes. Our pro-contrastive learning approach aims to bring samples from the same class closer together and samples from different classes farther apart. Furthermore, we specifically focus on separating harder negative class pairs. As a progressive step, both domains are adapted in this process.} 
    \label{model}
    \end{figure*}
\section{METHODOLOGY}

\mz{In this section, we will provide a detailed introduction to the GVG-PN framework. As depicted in Figure~\ref{model}, our model builds on the foundation of the GCN~\cite{luo2020progressive} and is divided into two distinct components: the domain-biased prototype model and the prototype-level contrastive learning. The feature extractor is used to extract relevant features from the input, which are then fed into the similarity graph generator to create relationship graphs. These graphs enable our model to generate aggregated features by considering both intra- and inter-domain samples.
Next, the source-biased and target-biased prototypes can be generated using these aggregated features, taking into account the intra- and inter-class relationships. In the final step, our pro-contrastive learning approach is utilized to focus on hard negative pairs in class-level, which helps our model to obtain discriminative features adaptively. This enables our model to generate features that are optimized for the final task.}

\subsection{Preliminaries}

\mz{Typically, an UDA dataset is defined as $D=\{D_{s}, D_{t}\}$, where
$D_{s}=\{(x_{s}^{i}, y_{s}^{i})\}_{i=1}^{n_{s}}$ represents the $n_{s}$ labeled samples in the source domain, while~$D_{t}=\{x_{t}^i\}_{i=1}^{n_{t}}$ denotes the ~$n_{t}$ unlabeled samples in the target domain. Here, $y_{s}^{i}\in\{1,2, \ldots, C\}$ denotes the label of ~$x_s^i$, $y_t$ represents the labels of the target domain samples, and both the source and target domains have the same $C$ categories. However, there is a significant difference in the distributions of the source and target domains, posing a challenge for UDA models to effectively train on the source domain and generalize well on the target domain. When the domain gap is within the UDA boundary, the model can achieve excellent performance. However, if the domain discrepancy is large, the model may collapse and fail to perform well on the target domain.}


\mz{Figure~\ref{model} represents a generic UDA framework that consists of two foundational networks. The first network is a feature extractor denoted $\mathcal{F}$ and parameterized by $\theta_{\mathcal{F}}$. This network extracts features from both the source and target domain, and the feature vector is denoted as $f =\mathcal{F}(x)$. The second network is a source classifier denoted $\mathcal{C}$ and parameterized by $\theta_{\mathcal{C}}$. This network is trained using labels from the source domain and produces classification predictions using the cross-entropy~(CE) loss function given by:}

\begin{equation}
\label{ce}
    \mathcal{L}_{c e} = -\frac{1}{{n}_{s}} \sum_{i=1}^{n_s} {y}_{s}^{i} \log (p(y | \mathcal{C}(\mathcal{F}({x}_{s}^{i})))),
\end{equation}
where $p(y | \mathcal{C}(\mathcal{F}(x_{s}^{i})))$ denotes the source classification predicted by the model.

\subsection{Domain-Biased Prototype Model}

\mz{Building on the GCN architecture, our proposed framework introduces a novel domain prototype generation scheme. Firstly, after feature extraction, features from different domains are mapped into the same subspace using GCN. In this subspace, intermediate domains can be generated using the aggregated features. Specifically, domain-biased prototypes can be obtained by using features from both the source and target domains.
These prototypes not only help reduce inherent differences between domains by aggregating sample features from another domain but also explore fine-grained clustering structures across domains through sample-level semantic similarity. Figure~\ref{model} provides a visualization of the proposed domain-biased prototype model, and a detailed description of the model can be found below.}

\subsubsection{Pipeline of GCN Layer}

\mz{Inspired by the GCN framework~\cite{luo2020progressive,kipf2016gcn}, we build a fully connected graph $G = (V, A)$ based on a training batch $B=\{B_{s}, B_{t}\}$.
In the beginning, the convolutional feature ${f}_{i}$ of each sample is used to represent the node $\mathbf{v}_{i} \in V$ in $G$.
The element ${a}_{i, j}$ in the affinity matrix $A$ denotes the semantic similarity score between node pairs $(\mathbf{v}_{i},\mathbf{v}_{j})$. Following~\cite{luo2020progressive}, the GCN $\mathcal{G}$ parameterized by $\theta_{\mathcal{G}}$ consists of three non-linear networks. The first is the affinity matrix generation layer $\mathcal{G}_{A}$, which updates the similarity scores of node pairs. The second is the node update layer $\mathcal{G}_{N}$, which aggregates features. The last is the graph classification layer $\mathcal{G}_{C}$, which produces $C$ outputs.}

For all node pairs $(\mathbf{v}_{i},\mathbf{v}_{j})$ in $B$, we compute their affinity score $\hat{a}_{i, j}^{(l)}$ at $l$-th layer to obtain the non-normalized affinity matrix $\hat{A}^{(l)}$:
\begin{equation}
    \label{aij}
    \hat{a}_{i, j}^{(l)}=\sigma(\mathcal{G}_{A}^{(l)}|\mathbf{v}_{i}^{(l-1)} - \mathbf{v}_{j}^{(l-1)}|),
\end{equation}
where $\sigma$ is a sigmoid function. The matrix $\hat{a}_{i, j}^{(l)}$ can be normalized to obtain the affinity matrix ${A}^{(l)}$ at the $l$-th layer:
\begin{equation}
    \label{A}
    {A}^{(l)}={D}^{-\frac{1}{2}}(\hat{A}^{(l)}+I) {D}^{-\frac{1}{2}},
\end{equation}
where $D$ is the degree matrix of $\hat{A}^{(l)}+I$ and $I$ is the identity matrix.

When the affinity matrix ${A}^{(l)}$ is obtained, all node features in $B$ can be updated at the $l_{th}$ layer:
\begin{equation}
    \label{v}    \mathbf{v}_{i}^{(l)}=\mathcal{G}_{N}^{(l)}( [ \mathbf{v}_{i}^{(l-1)},\sum_{j \in {B}} a_{i, j}^{(l)} \cdot \mathbf{v}_{j}^{(l-1)}  ] ),
\end{equation}
where $[\cdot, \cdot]$ is the connection operation. The output of $\mathcal{G}_{N}$ is the aggregated feature ${f}_{gcn}$.
Finally, ${f}_{gcn}$ is fed into the graph classification layer $\mathcal{G}_C$ and trained with cross-entropy loss $\mathcal{L}_{c e}^{gcn}$:
\begin{equation}
    \label{gcnce}
    \mathcal{L}_{c e}^{gcn}=-\frac{1}{B_{s}} \sum_{i \in {B}_s} y_{s}^{i} \log(p(y\mid \mathcal{G}_{C}({f}_{gcn}^{i}))  ).
\end{equation}


\subsubsection{Update of Affinity Matrix }
To investigate the semantic similarity relationship between pairs of samples, we construct the ground-truth matrix $T$ with labels to impose constraints on $\mathcal{G_A}$. The value of the element ${t_{i, j}}$ in $T$ is determined as follows:
\begin{equation}
    \label{tij}
    {t}_{i, j}=\left\{\begin{array}{ll}
     1, & \text { if } y_{i}=y_{j} \\
     0, & \text { otherwise }
\end{array}\right.,
\end{equation}
where $y$ represents the corresponding label of the sample. It is worth noting that for ${x_s} \in {B_s}$, the ground truth label ${y_s}$ from the source domain is used in the construction of matrix $T$. On the other hand, for ${x_t} \in {B_t}$, the corresponding pseudo-label $\hat{y_t}$ predicted by the source classifier is utilized in matrix $T$. 

Furthermore, in order to mitigate the impact of low-reliability samples, we introduce a threshold value denoted as $\delta$. 
If the predicted score of a sample ${x}_{t}$ is below the threshold  $\delta $, we mask off the edges associated with that sample in the unnormalized affinity matrix $\hat{A}$. This ensures that these edges are not optimized during training.

\zh{Unlike the fixed threshold used in the study~\cite{roy2021curriculum}, which overlooks low-confidence predictions in the early stages of training on the target domain, our approach adopts  an adaptive threshold. As the predictive capabilities of the network gradually improve during training, a fixed threshold may fail to effectively capture these changes, leading to a polarization in predictive performance. To address this issue, we employ an adaptive threshold determined by the mean and standard deviation of mini-batch samples. This adaptive approach enables the threshold to dynamically adjust, reflecting the evolving predictive capabilities of the network on the target domain.}

During the training phase, the model is optimized to align the output of the GCN with the ground-truth matrix $T$ constructed using the source classifier. The $\mathcal{G}_{A}$ is trained using binary cross-entropy loss, which is defined as follows:
\begin{equation}
    \label{bce}
    \mathcal{L}_{b c e} =\sum_{i \in B,j \in B} {t}_{i, j} \log p(\hat{a}_{i, j})+(1-{t}_{i, j}) \log (1-p(\hat{a}_{i, j})).
\end{equation}


\subsubsection{Aggregation of Domain-biased Prototype}
Most DA approaches aim to ensure semantic consistency by aligning the feature spaces of the source and target domains. Methods like MMD-based approaches~\cite{long2017deep,kang2019contrastive} and adversarial learning~\cite{long2018conditional,sharma2021instance} directly focus on aligning the global distribution but often ignore the fine-grained semantic relationships.
So these models may fail to adequately preserve the feature distribution during training.
Considering that prototypes are commonly used to represent category structures, we propose utilizing prototype alignment to transfer semantic knowledge from the source domain to the target domain. However, 
prototypes are typically calculated within a single domain, meaning they only capture information from one domain while disregarding the other. To address this, we leverage the affinity matrix ${A}$, constructed in both domains, to aggregate these prototypes and explore the cross-domain class-level distribution.
Specifically, the semantic similarity scores in ${A}$ serve as weights to facilitate the aggregation of features in our model. This allows the aggregated features to contain semantic information from both domains. Subsequently, domain-biased prototypes are generated based on these aggregated features, establishing a connection between the two domains through underlying fine-grained semantic relationships. This approach not only comprehensively describes the global distribution between domains but also reduces the discrepancy in the original prototype distributions.

Finally, the intermediate domain distributions are constructed with the domain-bias prototypes, which facilitates domain alignment. By leveraging the fine-grained semantic relationships captured by these prototypes, our approach benefits from improved alignment between the two domains.

Specifically, the domain-biased prototype refers to the class mean vector of aggregation features $f_{gcn}$ extracted from the respective domain. 
We calculate the source-biased $c_{sb}^{k}$ and target-biased prototype $c_{tb}^{k}$ for each category $k$ individually: 
\begin{equation}
\begin{aligned}
    \label{cs}
    c_{sb}^{k}=\frac{1}{\left|{D}_{s}^{k}\right|} \sum_{\left(x_{s}^{i}, y_{s}^{i}\right) \in {D}_{s}^{k}} f^s_{gcn}\\
    c_{tb}^{k}=\frac{1}{\left|{D}_{t}^{k}\right|} \sum_{\left(x_{t}^{i}, \hat{y}_{t}^{i}\right) \in {D}_{t}^{k}} f^t_{gcn}
\end{aligned}  ,
\end{equation}
where ${D}_{s}^{k}$ and ${D}_{t}^{k}$ denote the set of samples  with class $k$. Especially, in target domain, the prototypes are calculated with pseudo labels.

The update of the global class prototype follows the previous work~\cite{xie2018learning}, which coordinates the training process using an exponential moving average.
\begin{equation}
\begin{aligned}
    \label{css}
    c_{sb(I)}^{k} \leftarrow \rho c_{sb(I-1)}^{k}+(1-\rho) \hat{c}_{sb(I)}^{k}\\
    c_{tb(I)}^{k} \leftarrow \rho c_{tb(I-1)}^{k}+(1-\rho) \hat{c}_{tb(I)}^{k}
\end{aligned}  ,
\end{equation}
where $\rho$ denotes the trade-off parameters and it is set to 0.7 in all experiments. $\hat{c}_{sb(I)}^{k}$ and $\hat{c}_{tb(I)}^{k}$ are the class prototypes at the $I_{th}$ iteration.
With the iteration of the model, the more accurate prototypes can be obtained by aggregating the sample features in both the source and target domains.


\subsection{Prototype-Level Contrastive Learning}
In the UDA setting, discriminability is equally important as transferability for the downstream task. Simply aligning the intermediate domains globally is not sufficient. To explore the category structure more effectively between the intermediate domains, we implement a variant of contrastive learning at the prototype level.
More specifically, we introduce a pro-contrastive loss to enhance class-level discriminability, with a particular focus on the hard negative pairs. Furthermore, the loss function integrates the MMD loss to improve the discriminative features. By doing so, our model can achieve both transferability and discriminability, thereby improving overall performance.

\subsubsection{Preliminaries of InfoNCE}
Recently, most methods~\cite{he2020momentum,chen2020simple} have achieved superior performance in self-supervised learning using contrastive learning. The InfoNCE loss~\cite{oord2018representation} is a commonly used loss function that aims to minimize the distance between positive sample pairs while simultaneously maximizing the distance between negative sample pairs.
The InfoNCE loss is defined as:
\begin{small}
\begin{equation}
    \label{info}
    \resizebox{0.85\hsize}{!}{
    $\mathcal{L}_{\operatorname{InfoNCE}}\!=\!-\!\sum_{\boldsymbol{v}^{+} \in \mathcal{N}_{+}}\! \log\! \frac{\exp \left(\boldsymbol{v} \cdot \boldsymbol{v}^{+} / \tau\right)}{\exp \left(\boldsymbol{v} \cdot \boldsymbol{v}^{+} / \tau\right)\!+\!\sum\limits_{\boldsymbol{v}^{-} \in \mathcal{N}_{-}}\! \exp \left(\boldsymbol{v} \cdot \boldsymbol{v}^{-} / \tau\right)},$}
\end{equation}
\end{small}
where $\mathcal{N}_{+}$ and $\mathcal{N}_{-}$ denote the sets of positive and negative sample pairs related with $v$. 
The $\boldsymbol{v}$, $\boldsymbol{v}^{+}$ and $\boldsymbol{v}^{-}$ denote the $\ell_{2} \text {-normalized }$ features of the pair, respectively.

\subsubsection{Design of Pro-NCE}

To explore fine-grained semantic structures more effectively, we propose the use of prototype-level contrastive learning in a supervised manner. Traditional methods that treat all pairs equally are not appropriate since the similarity of class-level pairs can differ significantly.
To address the similarity imbalance existed in the class-wise problem, we reshape the standard class pairs to down-weight the loss assigned to easy pairs and up-weight the loss assigned to hard pairs. This is because hard negative class pairs tend to have higher similarity, while easy negative pairs tend to have lower similarity.
Specifically, we introduce a new loss function called ProNCE, which aims to increase the separation between harder negative pairs. Through the use of pro-contrastive loss, we can fully explore the fine-grained discriminability in semantic relations. In our experiments, we found that the best results were obtained using cosine distance as a metric $\phi$.
\begin{equation}
    \label{dist}
    \phi (u,v) = 1-\frac{u^{T} v}{\|u\|\|v\|},
\end{equation}
where a small value of $\phi$ represents high similarity, and vice versa.


For the $c_{th}$ prototype, the positive pair corresponds to the same prototype from the other domain, while the negative pairs are obtained from different categories in the two domains.
As mentioned earlier, our aim is to down-weight the loss assigned to easy pairs and up-weight the loss for hard pairs. To achieve this, we self-adaptively weight the prototype-level pairs, and rewrite the contrast loss as ProNCE, which leverages the metric $\phi$:
\begin{equation}
    \label{pronce}
    \resizebox{0.8\hsize}{!}{
    $\mathcal{L}_{\text {ProNCE}}= - \frac{1}{|C|} \sum_{c \in \mathcal{N}}\log \frac{\sum\limits_{c^{-} \in \mathcal{N}_{-}}  \exp(w{(c,c^{-})} \phi (c,c^{-})/ \tau)}{\exp( \phi (c,c^{+})/ \tau))},$}
\end{equation}
\noindent where $C$ refers to the total number of categories and $\mathcal{N}$ and $\mathcal{N}_{-}$ denote the sets of all pairs and negative pairs related to prototype $c$. It is important to note that for each prototype $c$, there is only one positive pair $c^+$ which has the same class from the different domain, while the other pairs $c^-$ are negative pairs obtained from both domains.
$\tau$ is a temperature hyper-parameter.
By using cosine similarity as the adaptively weighted function $w(i,j)$, our model can focus more on the negative pairs that are more similar, while reducing the effect of samples with low similarity.
\begin{equation}
  w(i,j)=cos(i,j),
\end{equation}
where $cos(i,j)$ denotes the cosine similarity function between two vectors $i$ and $j$.


\subsection{Overall Formulation}
According to TSA~\cite{li2021transferable}, in order to enhance the learning of high-level semantic features, we incorporate a mutual information loss into the model to improve the accuracy of the affinity matrix $A$. 
The formulation of the mutual information loss is as follows:
\begin{equation}
    \label{MI}
    \mathcal{L}_{M I}=\sum_{k=1}^{C} \hat{P}^{k} \log \hat{P}^{k}-\frac{1}{n_{t}} \sum_{i=1}^{n_{t}} \sum_{k=1}^{C} P_{t i}^{k} \log P_{t i}^{k} ,
\end{equation}
where $P_{t i}^{k}$ is the softmax outputs of target sample $x_t^i$ with class  $k$ and $\hat{\boldsymbol{P}}=\frac{1}{n_{t}} \sum_{j=1}^{n_{t}} \boldsymbol{P}_{t j}$.

In conclusion, the overall loss function of GVG-PN proposed in this paper is as follows:
\begin{equation}
    \label{GVG-PN}
    \mathcal{L}_{\text{GVG-PN}} = \mathcal{L}_{c e} + \lambda_1 \mathcal{L}_{c e}^{gcn} + \lambda_2 \mathcal{L}_{b c e}  + \lambda_3 \mathcal{L}_{M I}+ \gamma \mathcal{L}_{\text {ProNCE}} ,
\end{equation}

\noindent where $\lambda_1$, $\lambda_2$ and $\lambda_3$ represent trade-off parameters, while $\gamma$ is an adaptive parameter that increases with iterations. The ablation study conducted in our experiments could verify the contribution of each component.

Our GVG-PN contains the feature extractor $\mathcal{F}$, a classifier $\mathcal{C}$, a GCN $\mathcal{G}$ including the affinity matrix  $\mathcal{G}_{A}$, a node layer $\mathcal{G}_{N}$  and the graph classifier $\mathcal{G}_{C}$. The parameter $\theta_{\mathcal{F}}$, $\theta_{\mathcal{C}}$, $\theta_{\mathcal{G}}$ are optimized during the training process as follows:
\begin{equation}
    \begin{aligned}
\theta_{\mathcal{F}} \leftarrow \theta_{\mathcal{F}}-\eta \frac{\partial{\mathcal{L}_{\text{GVG-PN}}}}{\partial \theta_{\mathcal{F}}}, 
\\ \theta_{\mathcal{C}} \leftarrow \theta_{\mathcal{C}}-\eta \frac{\partial{\mathcal{L}_{\text{GVG-PN}}}}{\partial \theta_{\mathcal{C}}},
\\\theta_{\mathcal{G}} \leftarrow \theta_{\mathcal{G}}-\eta \frac{\partial{\mathcal{L}_{\text{GVG-PN}}}}{\partial \theta_{\mathcal{G}}},  
\end{aligned}
\end{equation}
where $\eta$ is the learning rate. The optimization procedure is following the basic CNN protocol.
In \textbf{Algorithm~\ref{alg}}, the pseudo-code for the proposed method GVG-PN is summarized.

\begin{figure}[!t]
\begin{algorithm}[H]
\caption{The GVG-PN Algorithm}
\label{alg}
\begin{algorithmic}[1]

\renewcommand{\algorithmicrequire}{\textbf{Input:}}
\REQUIRE \parbox[t]{\dimexpr\linewidth-\algorithmicindent}{Labeled source data $D_{s}=\{(x_{s}^{i}, y_{s}^{i})\}_{i=1}^{n_{s}}$.\\
Unlabeled target data $D_{t}=\{x_{t}^i\}_{i=1}^{n_{t}}$.}
\renewcommand{\algorithmicrequire}{\textbf{Require:}}
\REQUIRE Feature extractor $\mathcal{F}$, Source classifier $\mathcal{C}$ and GCN Network $\mathcal{G}$ (including affinity matrix generation layer $\mathcal{G}_A$, node update layer $\mathcal{G}_N$, and graph classifier $\mathcal{G}_C$).
\WHILE{not converge}
\STATE Utilize Eq.~(\ref{ce}) to establish $\mathcal{L}_{ce}$ for training $\mathcal{C}$.
\STATE Build the graph structure $G=(V,A)$, derive the elements $a_{i,j}$ in the affinity matrix $A$ using $\mathcal{G}_A$ in Eq.~(\ref{aij}) and (\ref{A}).
\STATE Derive the aggregated feature $f_{gcn}$ using $\mathcal{G}_N$ in Eq.(\ref{v}).
\STATE Utilize Eq.(\ref{gcnce}) to establish $\mathcal{L}^{gcn}_{ce}$ for training $\mathcal{G}_C$.
\STATE Build the ground-truth matrix $T$ using label information and establish $\mathcal{L}_{bce}$ using Eq.(\ref{bce}).
\STATE Utilize Eq.(\ref{cs}) and (\ref{css}) to calculate the domain-biased prototypes $c_{sb}$ and $c_{tb}$.
\STATE Utilize Eq.~(\ref{pronce}) to establish $\mathcal{L}_\text{ProNCE}$.
\STATE Utilize Eq.~(\ref{MI}) and (\ref{GVG-PN}) to compute $\mathcal{L}_\text{GVG-PN}$.
\STATE Utilize Eq.~(\ref{para}) to update the parameters $\theta_{\mathcal{F}}$, $\theta_{\mathcal{C}}$, and $\theta_{\mathcal{G}}$.
\ENDWHILE
\renewcommand{\algorithmicensure}{ \textbf{Output:}}
\ENSURE Predicted class of $x_{t}^i$.
\end{algorithmic}
\end{algorithm}
\end{figure}

\subsection{Theoretical Analysis}

In this section, we analyze our method and illustrate  how it improves the expected error boundary for target samples based on domain adaptation theory.

Firstly, we introduce the concept of $\alpha$-divergence between two distribution functions  $p(z)$ and $q(z)$, which can be defined as described in~\cite{aaa}:
\begin{equation}
\resizebox{0.85\hsize}{!}{
    $D_{\alpha}(p(z) \| q(z))=\frac{1}{\alpha(\alpha-1)}\left[\int p(z)^{\alpha} q(z)^{1-\alpha} d z-1\right] .$
    }
\end{equation}

 By considering the parameter $\alpha$ as an adjuster, the $\alpha$-divergence metric can smoothly transition between KL-divergence ($\alpha \rightarrow 1$) and reverse KL-divergence ($\alpha \rightarrow 0$) through the Hellinger distance ($\alpha \rightarrow 1/2$).
When $p(z) = q(z)$, the $\alpha$-divergence $D_{\alpha}(p(z) | q(z))$ equals zero.

According to \cite{it}, theoretical analysis based on $\alpha$-divergence in UDA methods can be provided to demonstrate the effectiveness of our approach.
Typically, we define the feature representation $z$ as the output of the feature extractor, and the classifier predicts the distribution $\hat{p}(y\mid z)$, which approximates the true distribution $p(y|z)$.
\begin{theo}
If $\alpha^{\prime} \in(0,1]$, define $\alpha=1-\alpha^{\prime}$ and assume that the loss $(-\log \hat{p}(y \mid z))$ is bounded by $M$, $y \in \mathcal{Y}$, $z \in \mathcal{Z}$ , then the result is:
\begin{equation}
\label{theory}
    \begin{aligned}
l_{\text {target }} & \leq l_{\text {source }}+\frac{M}{\sqrt{2}}\{\frac{1}{\alpha(\alpha-1) \log e}\}^{1 / 2} \\
& \times \sqrt{\log \{1-\alpha(1-\alpha) D_{\alpha}(q(z, y) \| p(z, y))\}}
    \end{aligned},
\end{equation}
where the loss of source domain is $ l_{\text {source }}=\mathbb{E}_{x, y \sim p(x, y), z \sim p(z \mid x)}[-\log \hat{p}(y \mid z)]$ and the loss of target domain is $l_{\text {target }}=\mathbb{E}_{x, y \sim q(x, y)}[-\log \hat{p}(y \mid x)]$.
\end{theo}
\emph{Proof:} From~\cite{ben2010theory} and \cite{long2018conditional}, Eq.~(\ref{theory}) can be rewritten as:
\begin{equation}
\label{r}
    l_{\text {target }} \leq l_{\text {source }}+\frac{M}{2} \int|p(z, y)-q(z, y)| d z d y,
\end{equation}
where the $p(z, y)$ and $q(z, y)$ represent the joint distributions of the source and target domains.
The absolute value is denoted by $|\cdot|$, and $\int|p(z, y)-q(z, y)| d z d y$ represents the total variation between the two distributions $p(z, y)$ and $q(z, y)$.

To calculate the upper bound of the target loss, we establish a relationship between the total variation and the $\alpha$-divergence by employing appropriate inequalities. Specifically, we establish a connection between the total variation and the $\text { Rényi }$ $\alpha \text {-divergence }$~($(R_{\alpha^{\prime}}(. \| .))$), which is closely related to the $\alpha \text {-divergence }$ mentioned in this paper. When $\alpha^{\prime} \in(0,1]$, this relationship can be expressed as given  in~\cite{rd}:
\begin{equation}
\label{r1}
\resizebox{0.89\hsize}{!}{
$\frac{\alpha^{\prime}}{2}(\int|p(z, y)-q(z, y)| d z d y)^{2} \log e \leq R_{\alpha^{\prime}}(p(z, y) \| q(z, y)).$
}
\end{equation}  

The $R_{\alpha^{\prime}}(p(z) \| q(z))$ is defined by $\frac{1}{\alpha^{\prime}-1} \log \int p(z)^{\alpha^{\prime}} q(z)^{1-\alpha^{\prime}} d z$. With the definition of $R_{\alpha}$, these two divergences are related by 
\begin{equation}
    \label{r2}
    \resizebox{0.89\hsize}{!}{$
        R_{\alpha^{\prime}}(p(z, y) \| q(z, y)) = 
        \frac{1}{\alpha^{\prime}-1} \log \{1-\alpha^{\prime}(1-\alpha^{\prime}) D_{\alpha^{\prime}}(p(z, y) \| q(z, y))\}.$
    }
\end{equation}
By inputting Eq.~(\ref{r2}) and Eq.(~\ref{r1}) into Eq.(~\ref{r}), the Eq.(~\ref{r}) can be  rewritten as:
\begin{equation}
    \begin{array}{l}
l_{\text {target }} \leq l_{\text {source }}+\frac{M}{\sqrt{2}}\{\frac{1}{\alpha^{\prime}(\alpha^{\prime}-1) \log e}\}^{1 / 2} \\
\quad \times \sqrt{\log \{1-\alpha^{\prime}(1-\alpha^{\prime}) D_{\alpha^{\prime}}(p(z, y) \| q(z, y))\}}
\end{array}.
\end{equation}

Finally, we change the variables as~$\alpha = \alpha^{\prime}$.
According to the  definition, $D_{\alpha^{\prime}}(p(z, y) \| q(z, y))=D_{1-\alpha^{\prime}}(q(z, y) \| p(z, y))$. This implies that by interchanging the positions of the distributions, the same value can be obtained when $\alpha^{\prime}$ is is replaced with $1 - \alpha^{\prime}$. 

\begin{mark1}
The above proof demonstrates that the loss function of the target domain has an upper bound. This bound is directly influenced by the classification loss function in the source domain and the discrepancy between the source and target distributions, as indicated by the $D_\alpha$ term.
Moreover, since $D_\alpha$ encompasses a range of divergence measures, it offers a more versatile parametric model for capturing distribution discrepancy. This enables a more flexible and comprehensive analysis of the differences between the domains.
\end{mark1}

\begin{mark1}
Based on Proposition 1, by gradually aligning the intermediate domain over iterations, $D_{\alpha}(q(z, y) \| p(z, y))$ diminishes, leading to a tighter bound on the second term.
Furthermore, the cross-entropy loss in the source domain is optimized integrated with the GCN classifier, minimizing the $l_{\text {source }}$. Simultaneously, the source domain sample space aggregates the target domain sample features, ndirectly minimizing $l_{\text {target }}$.
Under ideal conditions, when $D_\alpha \rightarrow 0$, the second term vanishes completely, indicating perfect alignment of the domain distributions. This implies that minimizing the target loss function is equivalent to minimizing the source loss function.
\end{mark1}

\begin{mark1}
Under extreme conditions, as $\alpha \rightarrow 1$ in Proposition 1, $D_{\alpha}(q(z, y) \| p(z, y)) \rightarrow D_{K L}(q(z, y) \| p(z, y))$. In this case, the $\text { L'Hopital's }$ rule can be applied.
\begin{equation}
    l_{\text {target }} \leq l_{\text {source }}+\frac{M}{\sqrt{2}} \sqrt{D_{K L}(q(z, y) \| p(z, y))}.
\end{equation}


Our approach leverages prototype-level contrastive learning to align the two domains effectively, enabling the learning of inter-domain semantic structures. By aligning the joint distributions $p(z, y)$ and $q(z, y)$, the second term is reduced, leading to a decrease in the upper bound of the target domain loss.
\end{mark1}

\section{EXPERIMENT}
\begin{table*}[]
\caption{Accuracy~(\%) of UDA on Office-31 using ResNet-50 as the backbone. The best performance is shown in \textbf{bold}.}
\centering
\setlength{\tabcolsep}{5mm}
\label{office}
\begin{tabular}{lccccccc}
\hline
Method     & A→W              & D→W                  & W→D             & A→D         & D→A        & W→A            & Avg           \\ \hline
Source-only  & 68.4             & 96.7                 & 99.3            & 68.9        & 62.5       & 60.7       & 76.1          \\
DANN~\cite{ganin2016domain}       & 82.0             & 96.9                 & 99.1            & 79.7        & 68.2       & 67.4       & 82.2          \\
MSTN~\cite{xie2018learning}       & 91.3             & 98.9                 & \textbf{100.0}  & 90.4        & 72.7       & 65.6       & 86.5          \\
CDAN+E~\cite{long2018conditional}     & 94.1             & 98.6                 & \textbf{100.0}  & 92.9        & 71.0       & 69.3       & 87.7          \\
CDAN+TFLGM~\cite{tflgm}              & 95.3             & 99.0                & \textbf{100.0}  & 94.1        & 73.2       & 73.3      & 89.2          \\
MEDM-LS~\cite{medm}    & 93.4         & 99.2        &99.8  & 93.2        & 75.1       & 75.4       & 89.3          \\
MCC+NWD~\cite{mcc}     & 95.5         & 98.7        &\textbf{100.0}  & 95.4        & 75.0       & 75.1       & 90.0          \\
SDAT+ELS~\cite{els}    & 93.6             & 99.0                 & \textbf{100.0}  & 93.4        & 78.7      & 77.5       & 90.4      \\
BIWAA~\cite{biwaa}    & 95.6             & 99.0                 & \textbf{100.0}  & 95.4        & 75.9      & 77.3       & 90.5      \\
BSP-TSA~\cite{li2021transferable}    & 96.0             & 98.7                 & \textbf{100.0}  & 95.4        & 76.7       & 76.8       & 90.6          \\
RSDA-MSTN~\cite{rsda}  & \textbf{96.1}    & \textbf{99.3}        & \textbf{100.0}  &95.8         & 77.4       & 78.9       & 91.1   \\
CO-HHDA~\cite{co-hhda}  & \textbf{96.1}    & 98.6       & \textbf{100.0}  &96.1        & 78.5       & 77.2       & 91.1   \\
FixBi~\cite{na2021fixbi}      & 96.1    & \textbf{99.3}        & \textbf{100.0}  & 95.0        & 78.7       & 79.4       & 91.4       \\ 
GSDE~\cite{gsde}      & \textbf{96.9}    & 98.8        & \textbf{100.0}  & \textbf{96.7}       & 78.3       & 79.2       & 91.7       \\ \hline
GVG-PN(Ours) & 95.7             & \textbf{99.3}        & \textbf{100.0}  &96.6 & \textbf{79.3}  & \textbf{79.6}  & \textbf{91.8} \\ \hline
\end{tabular}
\end{table*}

In this section, five  benchmark datasets of UDA are described firstly. Then the baseline methods and implementation details are introduced. Finally, we present extensive experimental results to demonstrate the effectiveness of our approach in comparison to the baseline methods.
\subsection{Datasets}
We evaluated our method on five public datasets, encompassing both small-scale and large-scale datasets. 

\textbf{Office-31}~\cite{saenko2010adapting}
is a well-established benchmark frequently used for DA tasks.
It comprises a total of 4,110 images, categorized into 31 different classes, and contains three distinct domains: Amazon~(A), Webcam~(W) and DSLR~(D). Following previous approaches\cite{tzeng2014deep,ganin2016domain}, we evaluate the adaptation performance across six different domain adaptation tasks: A $\rightarrow$ W, D $\rightarrow$ W, W $\rightarrow$ D, A $\rightarrow$ D, D $\rightarrow$ A and W $\rightarrow$ A.

\textbf{ImageCLEF-DA}\cite{caputoimageclef}
serves as the benchmark dataset for the ImageCLEF-DA 2014 domain adaptation challenge. It comprises three domains: Caltech-256~(C), ImageNet ILSVRC 2012~(I) and Pascal VOC 2012~(P).
Each domain consists of 12 categories, with 50 images per category.
The evaluation of transfer tasks on this dataset includes I $\rightarrow$ P, P $\rightarrow$ I, I $\rightarrow$ C, C $\rightarrow$ I, C $\rightarrow$ P and P $\rightarrow$ C.

\textbf{Office-Home}~\cite{venkateswara2017deep}
consists of a substantial collection of 15,500 images distributed across four domains, with each domain containing 65 different categories.
The four domains within this dataset are Art~(Ar), Clipart~(Cl), Product~(Pr) and Real-World~(Rw).
For our experiments, we evaluated all possible domain pairs, resulting in a total of 12 transfer tasks.

\textbf{VisDA-2017}~\cite{peng2017visda}
is a large-scale benchmark dataset for DA, comprising both a synthetic image domain and a real image domain.
It consists of a total of 12 categories. The synthetic image domain contains a vast collection of 152,409 images, while the real image domain comprises 55,400 samples sourced from MSCOCO~\cite{peng2017visda}.
For our evaluation, we utilized the synthetic images as the source domain and the real images as the target domain to train our model.

\textbf{DomainNet}~\cite{peng2019moment}
is one of the largest-scale datasets in DA, encompassing 345 categories with approximately 600,000 images. DomainNet consists of six domains with significant domain discrepancy: Clipart~(clp), Infograph~(inf), Painting~(pnt), Quickdraw~(qdr), Real~(rel) and Sketch~(skt).
Due to the large number of domains involved, we evaluate a total of 30 transfer tasks on this dataset.

\subsection{Baseline Methods \& Implementation Details}
\subsubsection{Baseline Methods}
\zh{In our experiments, we exclusively employed ResNet~\cite{he2016deep} as the backbone network. 
To ensure a fair comparison, we selected several classic and state-of-the-art approaches that utilized the same backbone network for benchmarking.
Therefore, we do not compare our method with those utilizing Transformer-based backbone networks.
On most datasets, excluding DomainNet, we compare our method against several baseline methods, including DANN~\cite{ganin2016domain}, CDAN~\cite{long2018conditional}, BSP-TSA~\cite{li2021transferable}, FixBi~\cite{na2021fixbi}, MSTN~\cite{xie2018learning}, RSDA-MSTN~\cite{rsda}, CRLP~\cite{crlp}, CO-HHDA~\cite{co-hhda}, TFLGM~\cite{tflgm}, MEDM-LS~\cite{medm}, GSDE~\cite{gsde}, BIWAA~\cite{biwaa}, ELS~\cite{els} and MCC~\cite{mcc}.
As for the DomainNet dataset, which is large and challenging, only a few algorithms have been verified on it. 
Therefore, for fairness, we compare our method with other algorithms that adopt CNN as the backbone network, including: MCD~\cite{saito2018maximum},CDAN~\cite{long2018conditional}, BNM~\cite{bnm}, SWD~\cite{swd}, and CGDM~\cite{cgdm}.
It is noteworthy that, unlike existing approaches~\cite{xie2018learning} that focus on both global and local alignments, our method does not directly align the target in the original domain.
Instead, we employ GCN to depict semantic similarity relationships among samples within batches, thereby generating domain-biased prototypes to characterize the intermediate domain-class structure. 
Gradually, the differences in the intermediate domain diminish, reducing the risk of distribution collapse.
Furthermore, aligning the original domain directly could lead to the complete misclassification of hard categories, where category prototypes reside in the wrong category space. 
To address this, we introduce weighted class-level adaptation, assigning greater separation weight to hard categories to displace their prototypes from the erroneous category space.
Distinct from other methods~\cite{na2021fixbi,inter} using an intermediate domain, which often rely on data augmentation techniques to generate the intermediate domain, we create the intermediate domain through the aggregation of sample features.
FixBi~\cite{na2021fixbi} represents a multi-stage approach, and our training process is end-to-end.
In contrast to methods like PGL~\cite{kipf2016gcn} and D-CGCT~\cite{luo2020progressive} that utilize GCN, we use GCN solely as a module for feature fusion. 
Additionally, due to differences in experimental setups, for fairness considerations, we refrain from direct comparisons with these two methods.
}

\subsubsection{Network Architecture}
To ensure fair comparison with baseline methods, we utilize ResNet-101 as the backbone network for the VisDA-2017 dataset and ResNet-50~\cite{he2016deep} pre-trained on ImageNet~\cite{russakovsky2015imagenet} as the
feature extractor $\mathcal{F}$  for all other datasets.
For the source classifier $\mathcal{C}$ and the GCN classifier $\mathcal{C}_{gcn}$, we employ fully connected layers for the classification task.
In the GCN network $\mathcal{G}$, the $\mathcal{G}_{A}$ component comprises two convolutional layers with $1\times1$ convolution kernels.
The structure of $\mathcal{G}_{N}$ is similar to $\mathcal{G}_{A}$, where the outputs maintain the same dimension as the input features.

\subsubsection{Implementation Details}
In this paper, our experiments are implemented with the PyTorch~\cite{paszke2019pytorch} framework.
\zh{
All experiments were conducted on a 24GB GeForce RTX 3090 GPU platform.
We performed training for 10,000 iterations on tasks from Office-31, ImageCLEF-DA, and Office-Home datasets, and for 20,000 iterations on tasks related to VisDA-2017 and DomainNet.
}
We employed stochastic gradient descent~(SGD) with a momentum of 0.9 for optimization.
\zh{The batchsize is set to 32. Through experimental adjustments, we set the initial learning rate $\eta$ to \text{1e-4} for the Office-31, ImageCLEF-DA, and VisDA-2017 datasets. For the Office-Home dataset, $\eta$ is configured at \text{5e-4}, and for the DomainNet dataset, it is set to \text{1e-3}.}
During training, the learning rate was adjusted using annealing arithmetic.
Regarding the update of the affinity matrix $A$, the construction of the ground-truth matrix $T$ depends on the pseudo-labels predicted by the source classifier. \zh{To obtain the accurate ground-truth matrix $T$, we compute the mean and standard deviation of the softmax probabilities corresponding to the highest predicted labels in each batch.  The confidence threshold $\delta$ is calculated across all mini-batches as $(mean - 2 \times std)$. In Eq.~(\ref{pronce}), we set the temperature hyperparameter $\tau$ to 0.05, following the configuration in work \cite{temperature}.}
Furthermore, the trade-off parameters in the total loss were set as $\lambda_1 = 0.3$, $\lambda_2 = 1$, and $\lambda_3 = 0.1$. The parameter $\gamma$ in the pro-contrast loss was set as $\gamma =\frac{2}{1+\exp (-\alpha p)}-1$, where $\alpha = 10$ and $p$ varies linearly within the range of \{0,1\} across the number of training iterations.

\begin{table*}[]

\caption{Accuracy~(\%) of UDA on ImageCLEF-DA using ResNet-50 as the backbone. The best performance is shown in \textbf{bold}.}
\centering
\label{clef}
\setlength{\tabcolsep}{5mm}
\begin{tabular}{lccccccc}
\hline
Method     & I→P           & P→I           & I→C           & C→I           & C→P           & P→C               & Avg           \\ \hline
Source-only  & 74.8      & 83.9      & 91.5      & 78.0      & 65.5      & 91.2          & 80.7          \\
DANN~\cite{ganin2016domain}       & 75.0      & 86.0      & 96.2      & 87.0      & 74.3      & 91.5          & 85.0          \\
CDAN~\cite{long2018conditional}       & 77.7      & 90.7      & 97.7      & 91.3      & 74.2      & 94.3          & 87.7          \\
MSTN~\cite{xie2018learning}       & 77.3      & 91.3      & 96.8      & 91.2      & 77.7      & 95.0          & 88.2          \\
CDAN+TFLGM~\cite{tflgm}         & 79.3      & 92.8      & 97.9      & 92.4      & 77.0      & 95.2          & 89.1          \\
MEDM-LS~\cite{medm}       & 78.2      & 93.3      & 97.2      & 93.0      & 78.3      & 95.5          & 89.3          \\
RSDA-MSTN~\cite{rsda}  & 79.8      & 94.5      & 98.0      & 94.2      & 79.2      & \textbf{97.3} & 90.5          \\
MCC+NWD~\cite{mcc}      & 79.8      & 94.5      & 98.0      & 94.2      & 80.0      & 97.5 & 90.7          \\
CRLP~\cite{crlp}       & 81.2          & 94.8          & 97            & \textbf{95.2} & 81.2         & 97.2              & 91.1         \\ \hline
GVG-PN(Ours) & \textbf{81.7} & \textbf{97.2} & \textbf{98.5} & 94.5          & \textbf{82.0} & 96.3              & \textbf{91.7} \\ \hline
\end{tabular}
\end{table*}

\begin{table*}[]
\caption{ Accuracy~(\%) of UDA on Office-home using ResNet-50 as the backbone. The best performance is shown in \textbf{bold}.}
\centering
\label{home}
\setlength{\tabcolsep}{1.5mm}
\begin{tabular}{lccccccccccccc}
\hline
Method     & Ar→Cl         & Ar→Pr         & Ar→Rw         & Cl→Ar         & Cl→Pr         & Cl→Rw         & Pr→Ar         & Pr→Cl         & Pr→Rw         & Rw→Ar         & Rw→Cl         & Rw→Pr         & Avg           \\ \hline
Source-only  & 34.9          & 50            & 58            & 37.4          & 41.9          & 46.2          & 38.5          & 31.2          & 60.4          & 53.9          & 41.2          & 59.9          & 46.1          \\
DANN~\cite{ganin2016domain}       & 45.6          & 59.3          & 70.1          & 47            & 58.5          & 60.9          & 46.1          & 43.7          & 68.5          & 63.2          & 51.8          & 76.8          & 57.6          \\
CDAN~\cite{long2018conditional}       & 49            & 69.3          & 74.5          & 54.4          & 66            & 68.4          & 55.6          & 48.3          & 75.9          & 68.4          & 55.4          & 80.5          & 63.8          \\
MSTN~\cite{xie2018learning}       & 40.8          & 70.3          & 76.3          & 60.4          & 68.5          & 69.6          & 61.4          & 48.9          & 75.7          & 70.9          & 55.0          & 81.1          & 65.7          \\
CDAN+TFLGM~\cite{tflgm}       & 51.4         & 72.0         & 77.2          & 61.7          &71.9         & 72.2          & 60.0          & 51.7          & 78.8          & 72.8          & 58.9          & 82.0         & 67.6          \\
RSDA-MSTN~\cite{rsda}  & 53.2          & 77.7          & 81.3          & 66.4          & 74.0          & 76.5          & 67.9          & 53.0          & 82.0          & 75.8          & 57.8          & 85.4          & 70.9          \\
BSP-TSA~\cite{li2021transferable}    & 57.6          & 75.8          & 80.7          & 64.3          & 76.3          & 75.1          & 66.7          & 55.7          & 81.2          & 75.7          & 61.9          & 83.8          & 71.2          \\
BIWAA~\cite{biwaa}    & 56.3          &78.4          & 81.2          & 68.0        & 74.5          & 75.7          & 67.9          & 56.1          & 81.2          & 75.2          & 60.1          & 83.8          & 71.5          \\
MEDM-LS~\cite{medm}       & 57.5       & 77.5          & 83.2            & 69.1          & 78.9          & 80.7            & 66.6          & 54.9          & 83.4          & 74.9          & 59.8          & 85.4            & 72.5          \\
CO-HHDA~\cite{co-hhda}       & 58.8         & 77.7         & 81.7            & 66.9          & 77.0          & 77.5            & 68.2          & 58.2          & 82.3          & 76.8         & 60.4          & 85.1            & 72.6         \\ 
MCC+NWD~\cite{mcc}       & 58.1         & 79.6        & 83.7            & 67.7         & 77.9          & 78.7            & 66.8          & 56.0          & 81.9          & 73.9         & 60.9          & 86.1            & 72.6          \\
SDAT+ELS~\cite{els}       & 58.2         & 79.7        & 82.5          & 67.5        & 77.2          & 77.2            & 64.6          & 57.9          & 82.2          & 75.4         & 63.1          & 85.5            & 72.6          \\
FixBi~\cite{na2021fixbi}      & 58.1          & 77.3          & 80.4          & 67.7          & \textbf{79.5}          &78.1          & 65.8          & \textbf{57.9}          & 81.7          & 76.4          & 62.9          & 86.7          & 72.7          \\
GSDE~\cite{gsde}      & 57.8          & 80.2         & 81.9         & 71.3         & 78.9          &80.5         & 67.4          & 57.2          & 84.0          & 76.1          & 62.5         & 85.7          & 73.6          \\ \hline    
GVG-PN(Ours) & \textbf{61.2} & \textbf{81.8} & \textbf{85.2} & \textbf{71.5} &79.4 & \textbf{81.2} & \textbf{70.0} &57.4 & \textbf{85.2} & \textbf{79.2} & \textbf{64.1} & \textbf{88.7} & \textbf{75.4} \\ \hline
\end{tabular}
\end{table*}

\begin{table}[]
\caption{ Accuracy~(\%) of UDA on VisDA-2017 using ResNet-101 as the backbone. The best performance is shown in \textbf{bold}.}
\centering
\label{vis}
\setlength{\tabcolsep}{5mm}
\begin{tabular}{lc}
\hline
Method     & Synthetic → Real \\ \hline
Source-only & 49.4             \\
DANN~\cite{ganin2016domain}       & 57.4             \\
MSTN~\cite{xie2018learning}       & 65.0             \\
CO-HHDA~\cite{co-hhda}   & 78.1             \\
RSDA-DANN~\cite{rsda}  & 79.1             \\
BSP-TSA~\cite{li2021transferable}    & 82.0    \\
MEDM-LS~\cite{medm}    & 82.4    \\
MCC+NWD~\cite{mcc}     & 83.7     \\ \hline
GVG-PN(Ours) & \textbf{83.9}    \\ \hline
\end{tabular}
\end{table}
\subsection{Comparison With Existing Methods}
\subsubsection{Results on Office-31}
Table~\ref{office} presents the classification accuracy for all tasks on the Office-31 dataset, highlighting the remarkable performance of our method, GVG-PN, with an average accuracy of 91.8\%.
Notably, in D $\rightarrow$ A task, our method even surpasses the state-of-the-art algorithm Fix-Bi~\cite{na2021fixbi} by an impressive margin of 0.6\%.
Compared with the similar works such as MSTN~\cite{xie2018learning} and RSDA-MSTN~\cite{rsda}, which also leverage prototype representation of semantic information, our method shows a substantial improvement of 5.3\% and 0.7\% in average accuracy, respectively. Additionally, our method outperforms the state-of-the-art methods GSDE~\cite{gsde} and FixBi~\cite{na2021fixbi} by 0.1\% and 0.4\%, respectively, further reflecting the reliability and effectiveness of our approach.

\subsubsection{Results on ImageCLEF-DA}
Table~\ref{clef} illustrates the performance of our method on the ImageCLEF-DA dataset. It is evident that our method achieves outstanding performance on most tasks. Particularly noteworthy is the P $\rightarrow$ I task, where we improve the accuracy by
2.7\% compared with RSDA-MSTN~\cite{rsda}. 
In comparison to the state-of-the-art method CRLP~\cite{crlp}, which can achieve 91.1\% accuracy on this dataset, our method GVG-PN still surpasses it by 0.6\%. This demonstrates the effectiveness of our model in alleviating model bias problems and exhibiting good generalization capabilities.

\subsubsection{Results on Office-Home}
The Office-Home dataset comprises four distinct domains, requiring evaluation across 12 domain adaptation tasks. As depicted in Table~\ref{home}, our  GVG-PN outperforms other baseline methods in 10 tasks when leveraging ResNet-50 as the backbone network. 
This dataset consists of 65 categories, and other DA methods that compute prototypes for each category separately may suffer from low discriminability due to the wide variety of categories. In contrast, our method leverages prototype learning to enable the model to learn both intra-class and inter-class relationships, resulting in more accurate information about the target domain categories.
On the Office-Home dataset, our method achieves an impressive average accuracy of 75.4\%, which is 4.5\% higher than the baseline method RSDA-MSTN and a 1.8\% improvement compared to the latest method GSDE~\cite{gsde}.

\subsubsection{Results on VisDA-2017}
VisDA-2017 is a challenging large-scaled dataset which has only two domains: synthetic domain and real domain. This dataset poses significant difficulties due to its vast number of images and the substantial gap between the domains.
The experimental results of our model on this dataset, using ResNet-101 as the backbone network, are presented in Table~\ref{vis}.
Notably, our method achieves a remarkable improvement of 18.9\% over the baseline method MSTN~\cite{xie2018learning} and a 0.2\% improvement over the state-of-the-art MCC+NWD~\cite{mcc} method.
These results demonstrate that our method not only performs well on classical datasets but also exhibits strong performance on datasets characterized by substantial domain gaps.

\begin{table*}[]
\caption{Accuracy~(\%) of UDA on DomainNet using ResNet-50 as the backbone. The best performance is shown in \textbf{bold}.}
\centering
\label{net}
\setlength{\tabcolsep}{0.7mm}
\begin{tabular}{c|ccccccc||c|ccccccc||c|ccccccc}
\hline
MCD~\cite{saito2018maximum}  & clp  & inf  & pnt  & qdr  & rel  & skt  & Avg. & CDAN~\cite{long2018conditional} & clp  & inf  & pnt  & qdr  & rel  & skt  & Avg. & BNM~\cite{bnm} & clp  & inf  & pnt  & qdr  & rel  & skt  & Avg. \\ \hline
clp  & -    & 15.4 & 25.5 & 3.3  & 44.6 & 31.2 & 24.0 & clp  & -    & 13.5 & 28.3 & 9.3  & 43.8 & 30.2 & 25.0 & clp  & -    & 12.1 & 33.1 & 6.2  & 50.8 & 40.2 & 28.5 \\
inf  & 24.1 & -    & 24.0 & 1.6  & 35.2 & 19.7 & 20.9 & inf  & 18.9 & -    & 21.4 & 1.9  & 36.3 & 21.3 & 20.0 & inf  & 26.6 & -    & 28.5 & 2.4  & 38.5 & 18.1 & 22.8 \\
pnt  & 31.1 & 14.8 & -    & 1.7  & 48.1 & 22.8 & 23.7 & pnt  & 29.6 & 14.4 & -    & 4.1  & 45.2 & 27.4 & 24.2 & pnt  & 39.9 & 12.2 & -    & 3.4  & 54.5 & 36.2 & 29.2 \\
qdr  & 8.5  & 2.1  & 4.6  & -    & 7.9  & 7.1  & 6.0  & qdr  & 11.8 & 1.2  & 4.0  & -    & 9.4  & 9.5  & 7.2  & qdr  & 17.8 & 1.0  & 3.6  & -    & 9.2  & 8.3  & 8.0  \\
rel  & 39.4 & 17.8 & 41.2 & 1.5  & -    & 25.2 & 25.0 & rel  & 36.4 & 18.3 & 40.9 & 3.4  & -    & 24.6 & 24.7 & rel  & 48.6 & 13.2 & 49.7 & 3.6  & -    & 33.9 & 29.8  \\
skt  & 37.3 & 12.6 & 27.2 & 4.1  & 34.5 & -    & 23.1 & skt  & 38.2 & 14.7 & 33.9 & 7.0  & 36.6 & -    & 26.1 & skt  & 54.9 & 12.8 & 42.3 & 5.4  & 51.3 & -    & 33.3\\
Avg. & 28.1 & 12.5 & 24.5 & 2.4  & 34.1 & 21.2 & 20.5 & Avg. & 27.0 & 12.4 & 25.7 & 5.1  & 34.3 & 22.6 & 21.2 & Avg. & 37.6 & 10.3 & 31.4 & 4.2  & 40.9 & 27.3 & 25.3\\ \hline
SWD~\cite{swd}  & clp  & inf  & pnt  & qdr  & rel  & skt  & Avg. & CGDM~\cite{cgdm} & clp  & inf  & pnt  & qdr  & rel  & skt  & Avg. & GVG-PN~(Ours) & clp  & inf  & pnt  & qdr  & rel  & skt  & Avg.\\ \hline
clp  & -    & 14.7 & 31.9 & 10.1 & 45.3 & 36.5 & 27.7 & clp  & -    & 16.9 & 35.3 & 10.8 & 53.5 & 36.9 & 30.7 & clp  & -    & 18.4 & 37.1 & 7.7 & 51.6 & 42.4 & \textbf{31.4} \\
inf  & 22.9 & -    & 24.2 & 2.5  & 33.2 & 21.3 & 20.0 & inf  & 27.8 & -    & 28.2 & 4.4  & 48.2 & 22.5 & \textbf{26.2} & inf  & 27.4 & -    & 30.1 & 2.8  & 41.1 & 22.7 & 24.8 \\
pnt  & 33.6 & 15.3 & -    & 4.4  & 46.1 & 30.7 & 26.0 & pnt  & 37.7 & 14.5 & -    & 4.6  & 59.4 & 33.5 & 30.0 & pnt  & 41.4 & 19.5 & -    & 4.0  & 53.4 & 37.8 & \textbf{31.2} \\
qdr  & 15.5 & 2.2  & 6.4  & -    & 11.1 & 10.2 & 9.1  & qdr  & 14.9 & 1.5  & 6.2  & -    & 10.9 & 10.2 & \textbf{8.7}  & qdr  & 12.1 & 2.9  & 5.7  & -    & 10.1 & 9.6 & 8.1\\
rel  & 41.2 & 18.1 & 44.2 & 4.6  & -    & 31.6 & 27.9 & rel  & 49.4 & 20.8 & 47.2 & 4.8  & -    & 38.2 & 32.0 & rel  & 50.2 & 23.4 & 50.6 & 3.6  & -    & 37.5 & \textbf{33.1} \\
skt  & 44.2 & 15.2 & 37.3 & 10.3 & 44.7 & -    & 30.3 & skt  & 50.1 & 16.5 & 43.7 & 11.1 & 55.6 & -    & 35.4 & skt  & 54.6 & 20.2 & 44.4 & 7.5 & 52.0 & -    & \textbf{35.7} \\
Avg. & 31.5 & 13.1 & 28.8 & 6.4  & 36.1 & 26.1 & 23.6 & Avg. & 36.0 & 14.0 & 32.1 & \textbf{7.1}  & \textbf{45.5} & 28.3 & 27.2 & Avg. & \textbf{37.1} & \textbf{16.9} & \textbf{33.6} & 5.1  & 41.6 & \textbf{30.0} & \textbf{27.4}\\ \hline
\end{tabular}
\end{table*}
\subsubsection{Results on DomainNet}
DomainNet is an extensive dataset consisting of hundreds of image categories. Recently, state-of-the-art DA methods have adopted the Transformer architecture and achieved impressive results. However, in our experiments, we chose to use the CNN-based ResNet-50 framework for feature extraction due to its specificity and effectiveness.
Table~\ref{net} presents a comparison of several methods utilizing the same backbone network.
Notably, our proposed GVG-PN method achieves the highest average accuracy in 8 out of 12 cases and ultimately attains the highest accuracy of 27.4\% in average.
These results highlight the strong generalization capabilities of our approach in tackling the challenges posed by the DomainNet dataset.

\section{Discussion}
\begin{table*}[]
\centering
\begin{threeparttable}[t]
\caption{Ablation analysis (\%) on Office-31. The best performance is shown in \textbf{bold}.}
\label{abla}
\renewcommand\arraystretch{1.2}
\begin{tabular}{c|c|c|c|c||c|c|c|c|c|c|c}
\hline
 $\mathcal{L}_{c e}$ &$\mathcal{L}_{M I}$  & $\mathcal{L}^{g c n}_{c e}$ & $\mathcal{L}_{b c e}$ &$\mathcal{L}_{\text {ProNCE}}$ &A→W &D→W &W→D &A→D  & D→A & W→A & Avg  \\ \hline 
$\checkmark$ &              &   &      &    & 68.4      & 96.7              & 99.3             & 68.9              & 62.5 & 60.7 & 76.1 \\ 
$\checkmark$ & $\checkmark$ &  &      &    & 70.6            & 97.4              & 97.8              & 79.7      & 60.1 & 56.5 & 77.0 \\ 
$\checkmark$ & $\checkmark$ & & $\checkmark$     &    & 92.3   & 98.8    & 100.0 & 95.1  & 73.7  &73.5 & 89.2\\ 
$\checkmark$ & $\checkmark$ &       &    &$\checkmark$ & 89.8& 98.7 & 100.0 & 95.9& 73.8 & 71.9& 88.4\\ 
$\checkmark$ & $\checkmark$ & $\checkmark$ &      &    & 81.5& 96.4  & 98.5  & 89.3      & 66.5 & 65.4 & 82.9 \\ 
$\checkmark$ & $\checkmark$ & $\checkmark$      & $\checkmark$   & & 93.8& 99.2 & 100.0 & 95.6& 76.5 & 77.7& 90.5\\ 
$\checkmark$ & $\checkmark$ & $\checkmark$    &    &$\checkmark$   & 91.9& 99.1 & 100.0 &95.0& 74.7 & 74.1& 89.1\\

\hline
 &$\checkmark$&$\checkmark$&$\checkmark$&$\checkmark$  &68.4\tnote{1} &95.7\tnote{1} & 91.7\tnote{1} & 67.1\tnote{1}&63.2\tnote{1} & 59.6\tnote{1}&73.4\tnote{1} \\ 
$\checkmark$ & $\checkmark$ & $\checkmark$    & $\checkmark$   &$\checkmark$   & 93.3\tnote{2}& 98.9\tnote{2} & 100.0\tnote{2} & 95.0\tnote{2}& 75.2\tnote{2} & 74.5\tnote{2}& 89.5\tnote{2}\\
$\checkmark$ & $\checkmark$ & $\checkmark$    & $\checkmark$   &$\checkmark$   & \textbf{95.7}& \textbf{99.3} & \textbf{100.0} & \textbf{96.6}& \textbf{79.3} & \textbf{79.6}& \textbf{91.8}\\
\hline
\end{tabular}
\begin{tablenotes}
       \item [1] The $T$ in $\mathcal{L}_{b c e}$ is constructed  with the pseudo-labels predicted by $\mathcal{G}_C$.
       \item [2] The results are  from the  classifier $\mathcal{C}$. More experiments are presented in Table~\ref{class}.
     \end{tablenotes}
\end{threeparttable}
\end{table*}

\begin{table*}[]
\caption{The results(\%) of different classifiers on three datasets.The best performance is shown in \textbf{bold}.}
\centering
\label{class}
\begin{tabular}{c|c|c|c|c|c|c|c|c|c}
\hline
Datasets & \multicolumn{3}{c|}{Office-31} & \multicolumn{3}{c|}{Office-Home}  &\multicolumn{3}{c}{ImageCLEF-DA}\\ \hline
  Task &D→W &D→A&W→A&Rw→Ar &Pr→Rw & Ar→Pr&I→P &C→I  &P→C   \\ \hline
  Classifier~$\mathcal{C}$&98.9& 75.2&74.5& 72.5 &78.1&76.7 &79.5&91.6&95.3 \\
  Classifier~$\mathcal{G}_{C}$&\textbf{99.3}&\textbf{79.3}&\textbf{79.6}&\textbf{78.2}&\textbf{84.2}&\textbf{80.2}&\textbf{81.7}&\textbf{94.5}&\textbf{96.3} \\ \hline
  
\end{tabular}
\end{table*}

In our framework, we mainly focus on two key aspects: the construction of domain-biased prototypes to capture global semantic information and prototype-level contrastive learning to enhance local relationships.
To provide a more comprehensive understanding of our proposed approach, we conducted ablation studies and discussions on three datasets. These studies allowed us to analyze the individual contributions of different components and further validate the effectiveness of our method.

\subsubsection{Ablation study}
We perform an ablation study on the Office-31 dataset to evaluate the effectiveness of each module in our proposed GVG-PN framework.
We propose a method that comprises two essential components: the generation of domain-biased features and prototype-level contrastive learning. In order to evaluate the efficacy of each component, we conducted an ablation analysis under different baselines, examining the results obtained from various model variants with specific loss functions removed.

In Table~\ref{abla}, the first row represents the performance when only utilizing the source classifier cross-entropy loss, which serves as a low baseline with an average recognition performance of 76.1\% on the target domain. 
To further optimize the model outputs, we subsequently incorporate mutual information loss on the target domain. This addition yields a slight improvement in prediction performance, reaching 77.0\%, which can be regarded as another baseline.
Furthermore, a better model has the capability to provide more precise graph relationships, thereby facilitating the exploration of manifold structures.

Based on the two baselines, we evaluated the performance of our proposed loss function in the third and fourth rows. 
In the third row, we introduced the $\mathcal{L}_{b c e}$ loss to preserve the semantic relations between samples. This allowed us to obtain domain-biased prototypes by aggregating features from the cross domains. With the $\mathcal{L}_{b c e}$ loss, the model achieved an average performance of 89.2\%. These results highlight the value of our semantic structures in modeling sample similarity relations across domains and exploring category information.
Furthermore, we analyzed the discriminability among semantic classes based on the obtained prototypes. We utilized the
$\mathcal{L}_{\text {ProNCE}}$ loss to optimize the prototypes, ensuring that prototypes from the same category but different domains are brought closer together, while prototypes from different categories, particularly the challenging hard prototype pairs, are separated further apart. Leveraging the pro-contrastive loss, our model achieved a performance of 88.4\% based on $\mathcal{L}_\text {ProNCE}$ in the fourth row. Notably, we observed remarkable improvements of 
95.9\% and 73.8\% on the A $\rightarrow$ D and D $\rightarrow$ A tasks, respectively.

In this paper, we select to utilize the classification outputs from the GCN classifier, as it has demonstrated strong performance. To evaluate the effectiveness of this approach, we present the baseline results in the fifth row, achieving an accuracy of 82.9\%. Furthermore, we evaluate the efficacy of the two losses and present the corresponding results in Table~\ref{abla} as 90.5\% and 89.1\%. 
Significantly, the results obtained using the $\mathcal{L}_{b c e}$ loss appear to be superior to those obtained using the $\mathcal{L}_{\text {ProNCE}}$ loss. One possible reason for this discrepancy could be the absence of constraints on manifold structures, leading to unreliable generated prototypes.

Additionally, in the $\mathcal{L}_{b c e}$ loss, the ground truth matrix $T$ is constructed with the pseudo labels obtained from the source classifier $\mathcal{C}$. To verify the effectiveness of this strategy, we also present the results obtained using pseudo-labels generated with
$\mathcal{G}_C$. 
The results displayed in row eight demonstrate the effectiveness of this strategy. The reason behind this lies in the fact that when unreliable pseudo-labels are predicted using the initial GCN model, it is possible for the labels to collapse the GCN model during the iterative process.

Moreover, Table~\ref{class} and the last rows in Table~\ref{abla}  present the results predicted by different classifiers with different datasets. The term Classifier~$\mathcal{G}_C$ refers to the graph classifier, while Classifier~$\mathcal{C}$ represents the source classifier. Obviously,
the predictions made by Classifier~$\mathcal{G}_C$ outperform those of Classifier~$\mathcal{C}$. These results demonstrate the effectiveness of our output strategy.

In conclusion, our progressive alignment framework effectively constrains inter-domain semantic information and enhances the discriminability of the model, resulting in improved performance.

\subsubsection{ Effectiveness of ProNCE and domain-biased prototypes}
\begin{table}[]
\caption{Comparison of three different loss functions and two different prototype calculation methods on tasks D→A and W→A.}
\centering
\label{adva}
\renewcommand\arraystretch{1.3} 
\begin{tabular}{c|cc||cc}
\hline
\multirow{2}{*}{Type}   & \multicolumn{2}{c||}{w/o domain-biased}                             & \multicolumn{2}{c}{ domain-biased}                              \\ \cline{2-5} 
                        & \multicolumn{1}{c|}{D→A} & W→A & \multicolumn{1}{c|}{D→A} & W→A \\ \hline
$\mathcal{L}_{\text{SM}}$     & \multicolumn{1}{c|}{72.7}                  &65.6                   & \multicolumn{1}{c|}{77.2}              & 75.3              \\ \hline
$\mathcal{L}_{\text{InfoNCE}}$ & \multicolumn{1}{c|}{75.6}                  &72.3                   & \multicolumn{1}{c|}{77.4}                  &73.9                   \\ \hline
$\mathcal{L}_{\text {ProNCE}}$(Ours)  & \multicolumn{1}{c|}{77.8}              & 74.5              & \multicolumn{1}{c|}{\textbf{78.8}}              & \textbf{76.5}              \\ \hline
\end{tabular}
\end{table}


\begin{figure*}[t]
    \centering
    \subfloat[\fontsize{4pt}{8pt}\selectfont]{\includegraphics[width=0.24\textwidth]{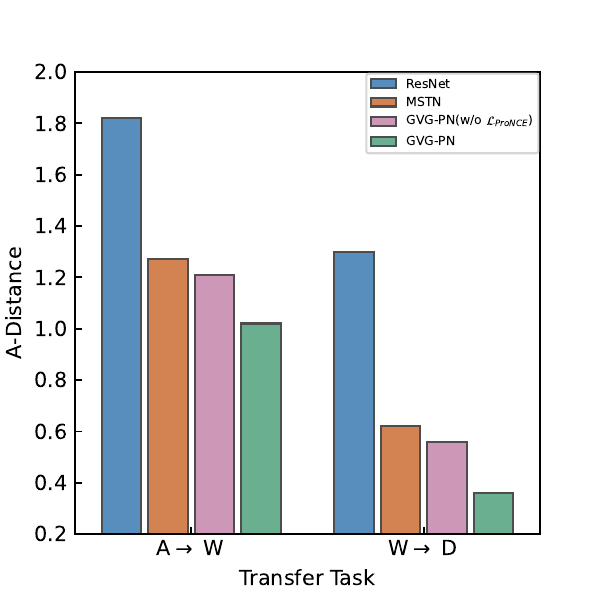}}
    \hfill
    \subfloat[\fontsize{4pt}{8pt}\selectfont]{\includegraphics[width=0.24\textwidth]{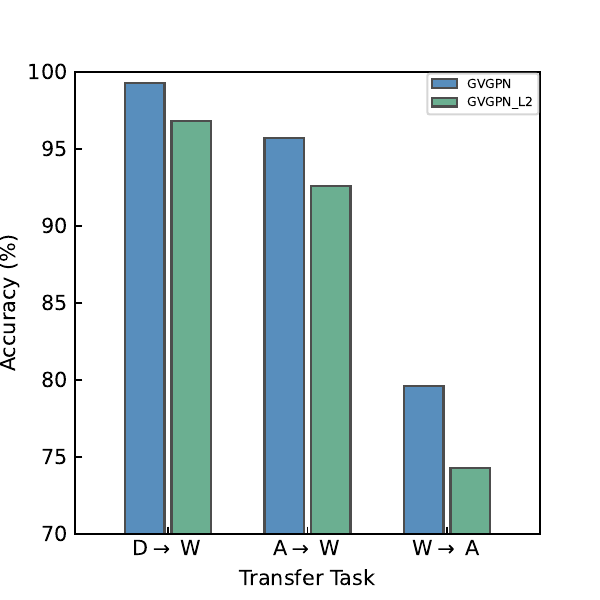}}
    \hfill
    \subfloat[\fontsize{4pt}{8pt}\selectfont]{\includegraphics[width=0.24\textwidth]{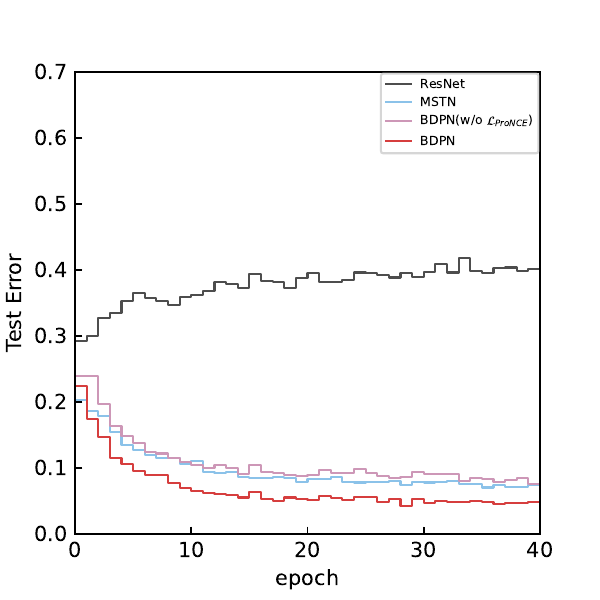}}
    \hfill
    \subfloat[\fontsize{4pt}{8pt}\selectfont]{\includegraphics[width=0.24\textwidth]{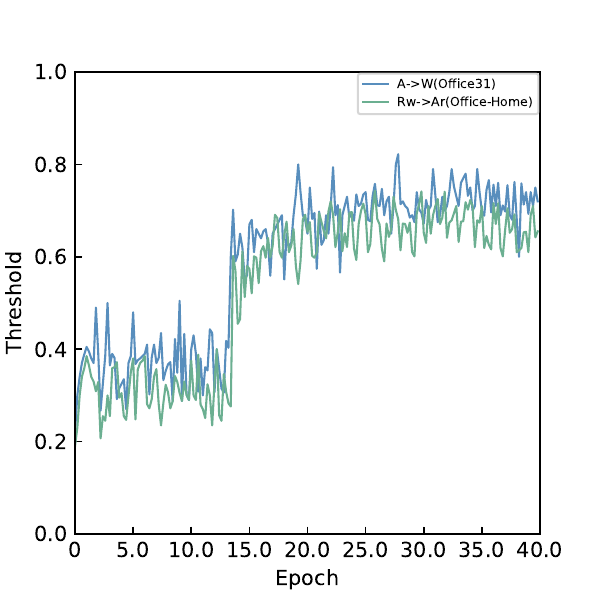}}
    \caption{
Discussion of various model analyses:~(a) Quantitative distribution differences between domains measured using $\mathcal{A}$-distance following domain adaptation.~(b) Accuracy comparison of GVG-PN and the variant of GVG-PN on three tasks on the Office31 dataset.~(c) Convergence of test errors among different models.~(d) During the training process, dynamic threshold adaptive changes are observed on tasks A $\rightarrow$ W and Rw $\rightarrow$ Ar.
}
    \label{ddcs}
\end{figure*}

In this part of the experiment, we removed $\mathcal{L}_{M I}$ from the framework in order to make a pure comparison of the experimental results.

To highlight the advantages of $\mathcal{L}_{\text {ProNCE}}$, we incorporated two alternative loss functions in our framework. 
Firstly, we used the semantic matching loss, $\mathcal{L}_{\text{SM}} = 1-\frac{c_{st}^{T} c_{ts}}{\|c_{st}\|\|c_{ts}\|}$~\cite{rsda}, from MSTN~\cite{xie2018learning},  which aims to reduce the distance between the prototypes of the same category across domains to achieve domain alignment.
Secondly, we replaced it with InfoNCE in Eq~(\ref{info}), which selects positive and negative sample pairs for training according to our strategy and  aims to keep the prototypes of the same category close and those of different categories far apart.
Table~\ref{adva} presents the performance under different loss functions for two challenging tasks,
D $\rightarrow$ A and W $\rightarrow$ A. As the loss functions are based on prototypes, we evaluate their effectiveness in two experimental settings: without or with the domain-biased prototypes.
Without using the domain-biased prototypes~(w/o domain-biased), $\mathcal{L}_{\text {ProNCE}}$ achieves a performance of 77.8\%  and 74.5\%, respectively, showcasing significant improvements compared to the other two losses. When employing the domain-biased prototype strategy~(domain-biased), our ProNCE still outperforms the best of the other two losses by 1.4\% and 1.2\%, respectively. These results demonstrate the effectiveness and superiority of $\mathcal{L}_{\text{ProNCE}}$ in improving DA performance.

Next, we demonstrate the effectiveness of our strategy in utilizing domain-biased prototypes.
In the experiments, 'w/o domain-biased' refers to directly calculating prototypes using the output features without the process of feature aggregation, while 'domain-biased' represents our method that incorporates the domain-biased prototype strategy. 
As shown in Table~\ref{adva}, we observe significant performance improvements for every task in the 'domain-biased' setting. 
Specifically, when employing  $\mathcal{L}_{\text {ProNCE}}$ in the 'domain-biased' approach, we achieve a performance improvement of 1.0\% and 2.0\% on the D $\rightarrow$ A and W $\rightarrow$ A tasks, respectively.
This result demonstrates that utilizing the feature space of intermediate domains for progressive alignment can effectively mitigate the challenges associated with large domain discrepancies, leading to improved adaptation performance.


\begin{figure}[t]
    \centering
    \subfloat[\fontsize{4pt}{4pt}\selectfont Non-adaptation]{\includegraphics[width=0.11\textwidth ]{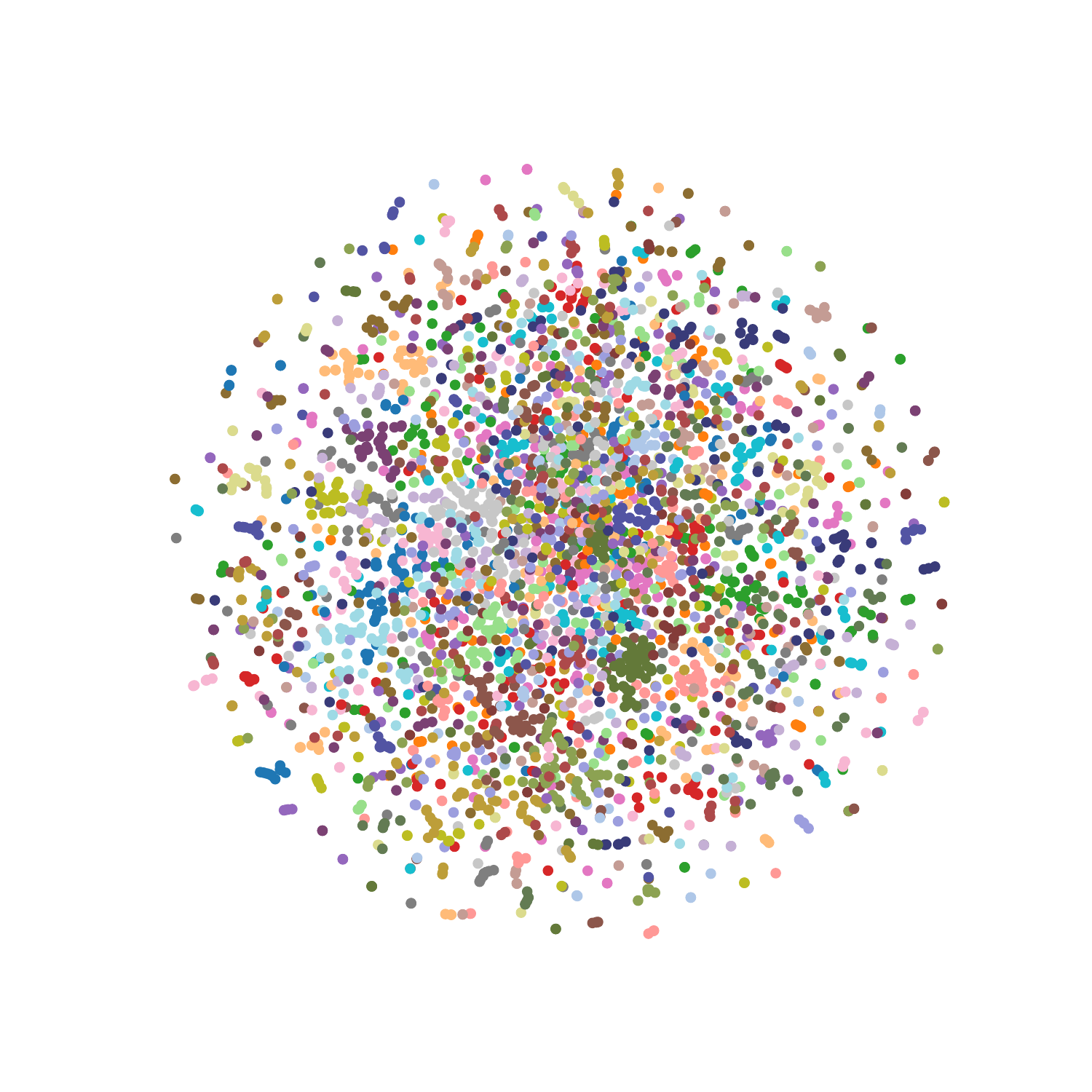}}
    \hfill
    \subfloat[\fontsize{4pt}{4pt}\selectfont MSTN]{\includegraphics[width=0.11\textwidth]{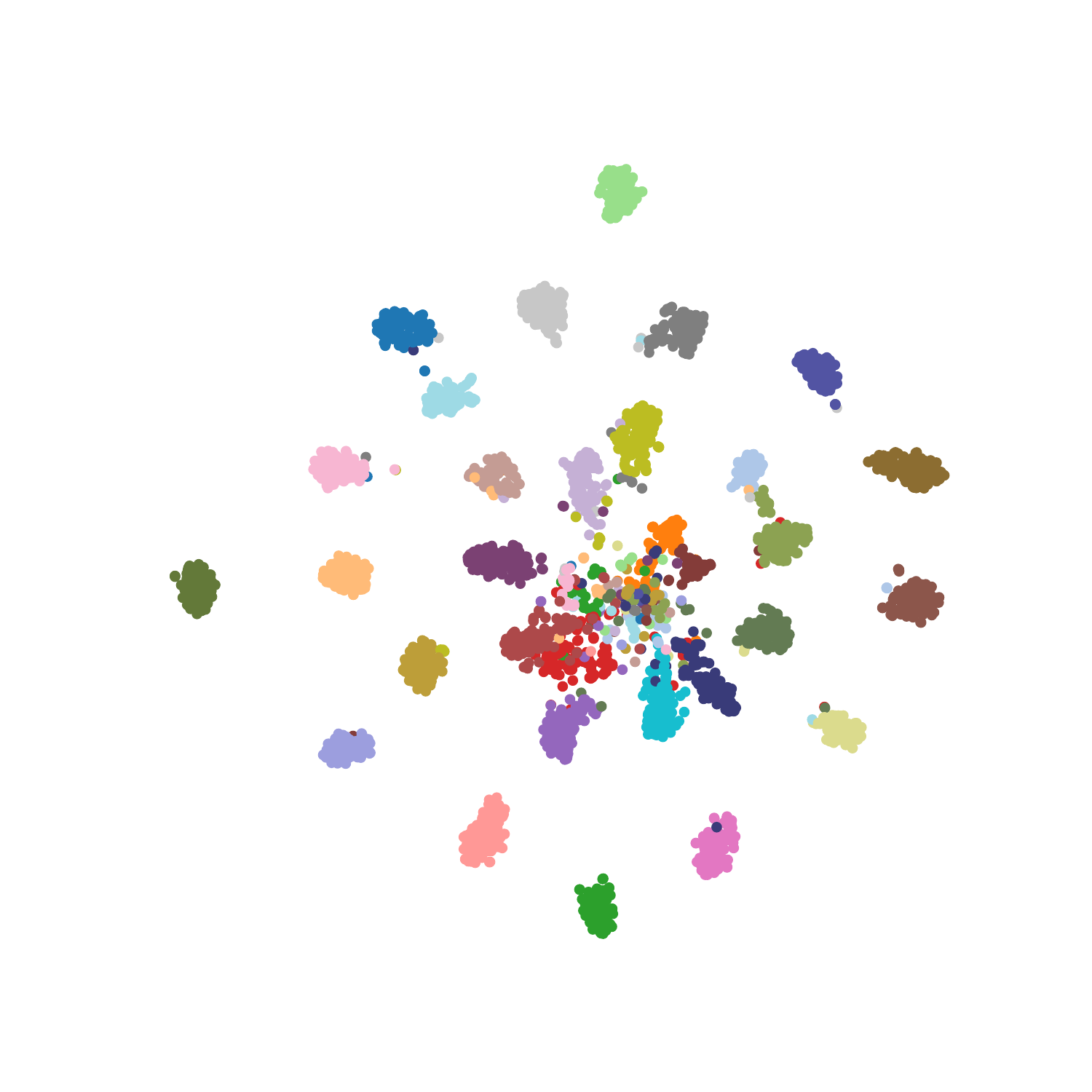}}
    \hfill
    \subfloat[\fontsize{4pt}{4pt}\selectfont GVG-PN(w/o $\mathcal{L}_{\text {ProNCE}}$)]{\includegraphics[width=0.11\textwidth]{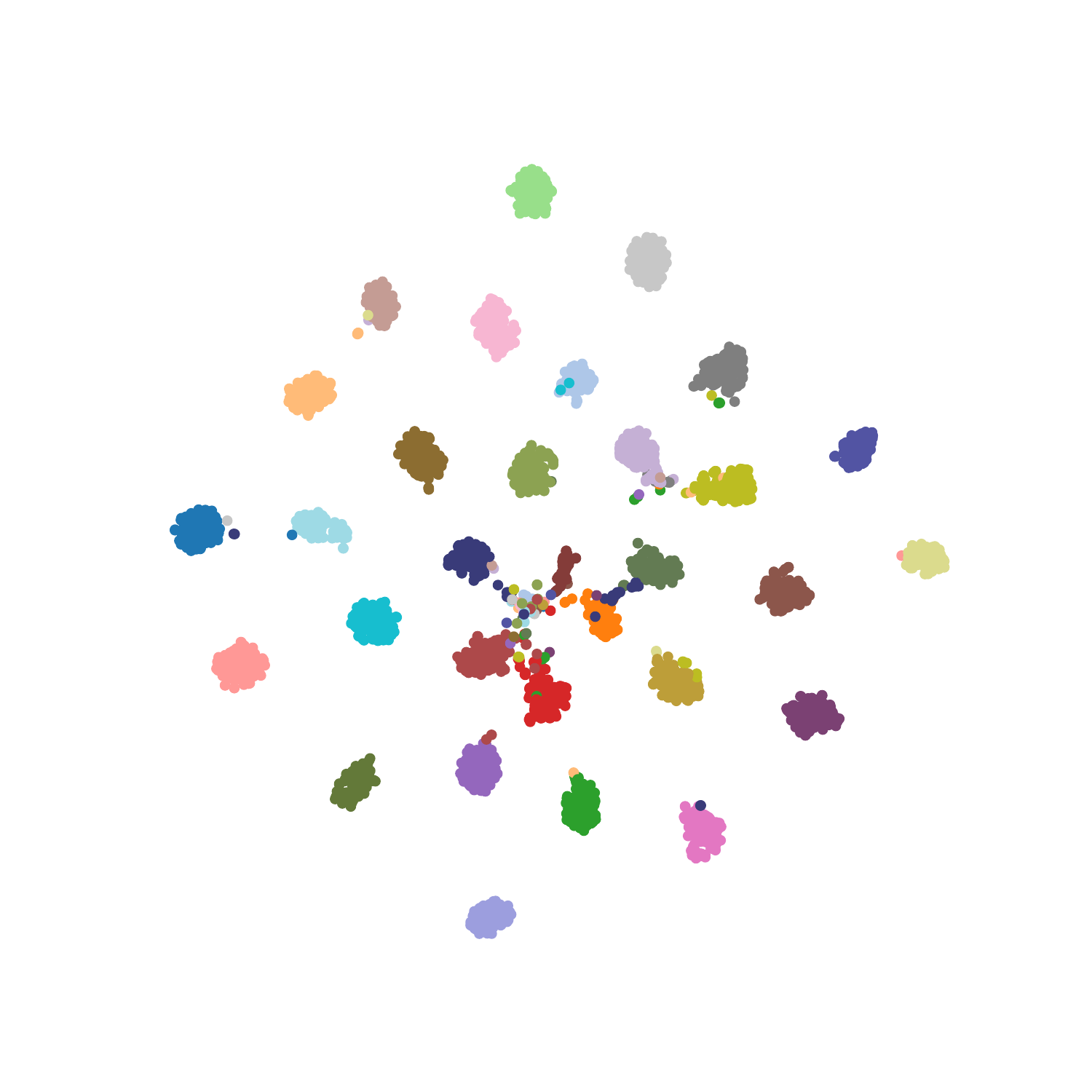}}
    \hfill
    \subfloat[\fontsize{4pt}{4pt}\selectfont GVG-PN]{\includegraphics[width=0.11\textwidth]{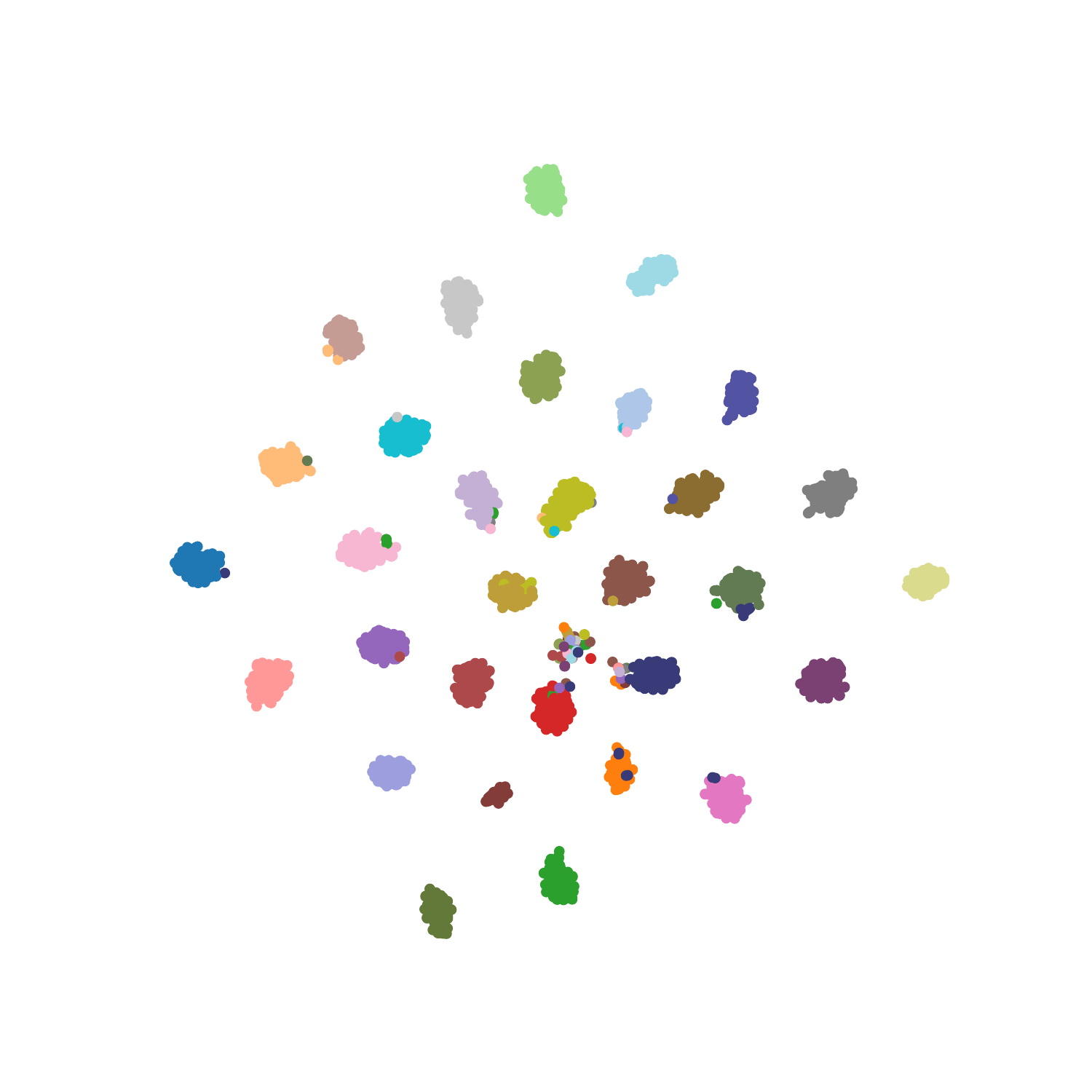}}

    \subfloat[\fontsize{4pt}{4pt}\selectfont Non-adaptation]{\includegraphics[width=0.11\textwidth ]{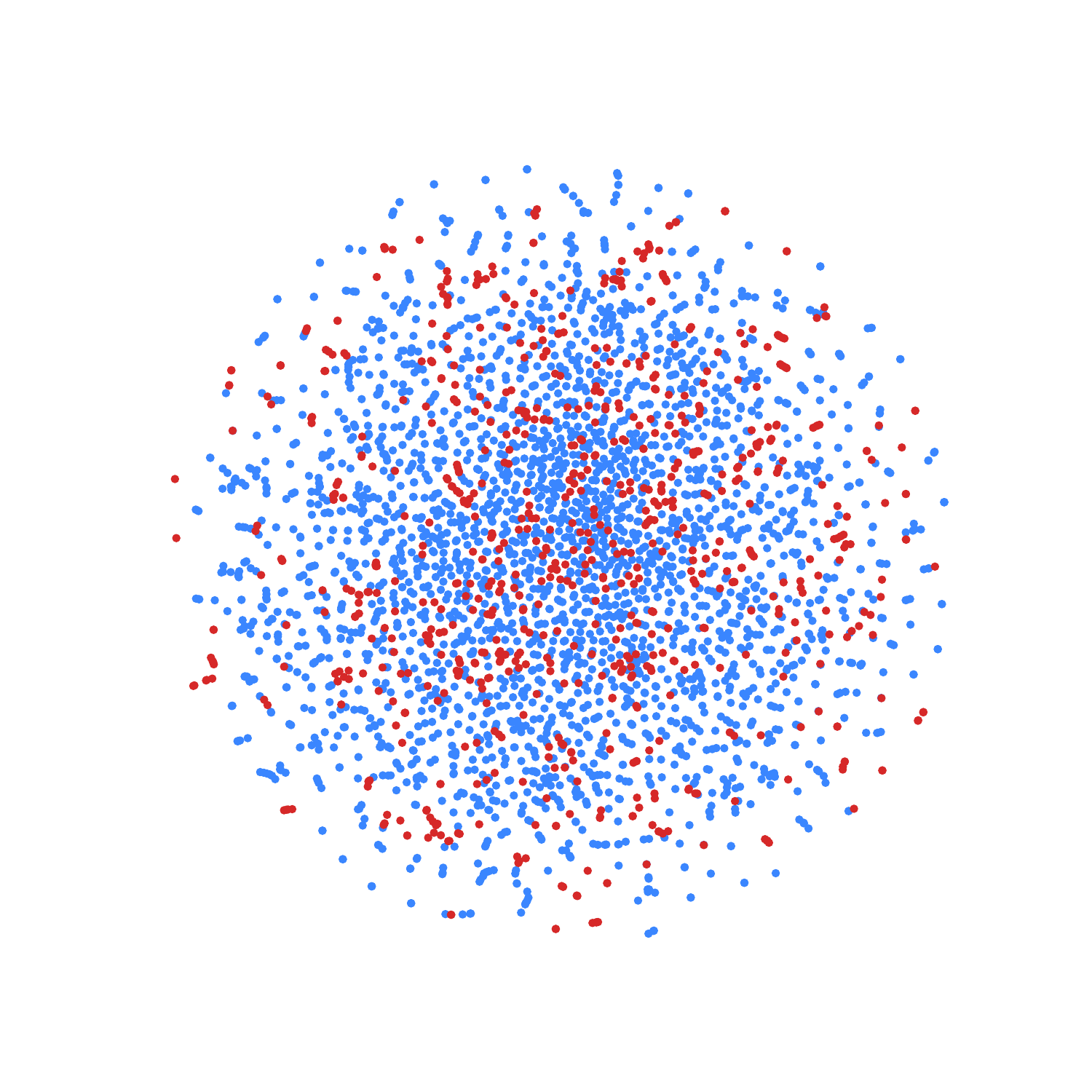}}
    \hfill
    \subfloat[\fontsize{4pt}{4pt}\selectfont MSTN]{\includegraphics[width=0.11\textwidth]{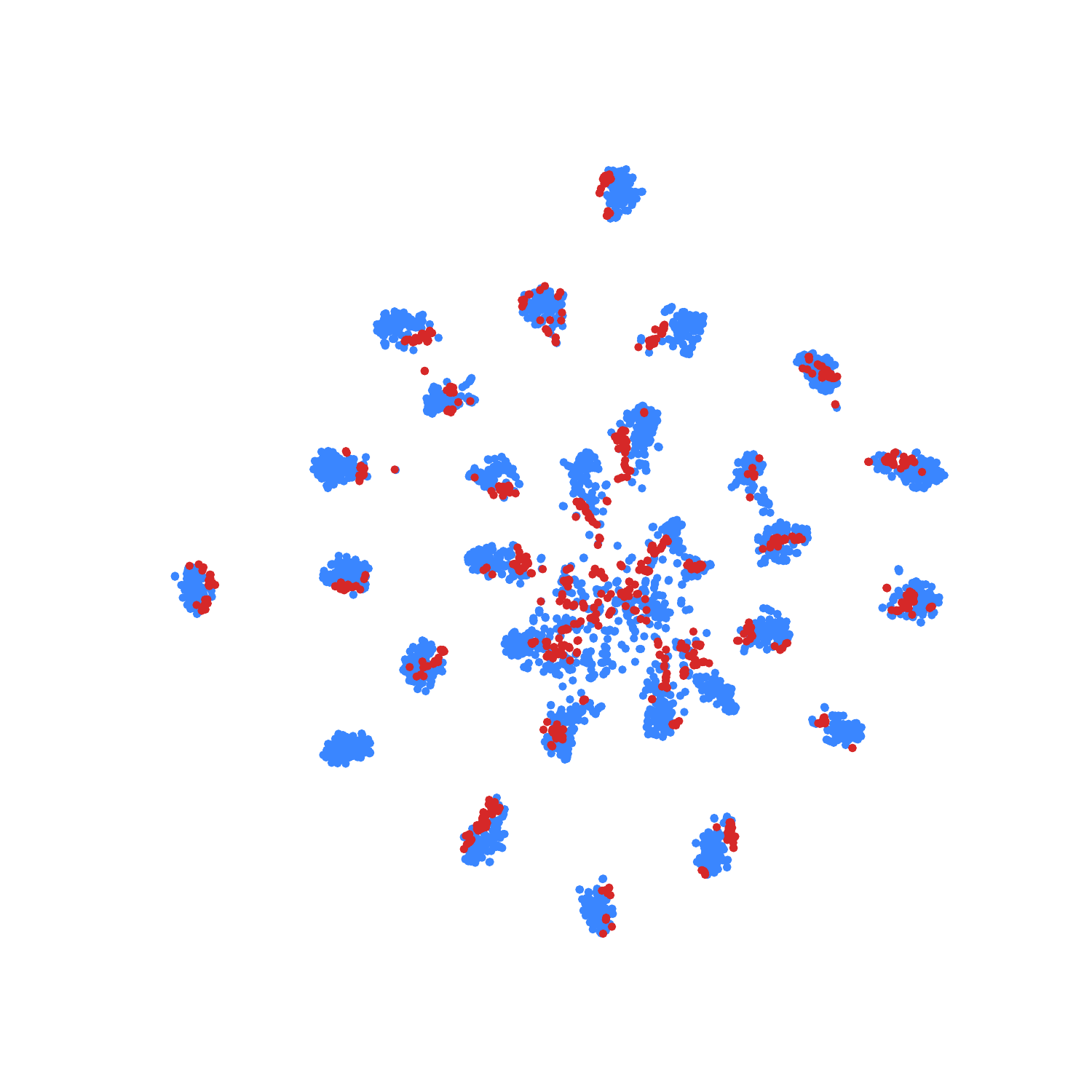}}
    \hfill
    \subfloat[\fontsize{4pt}{4pt}\selectfont GVG-PN(w/o $\mathcal{L}_{\text {ProNCE}}$)]{\includegraphics[width=0.11\textwidth]{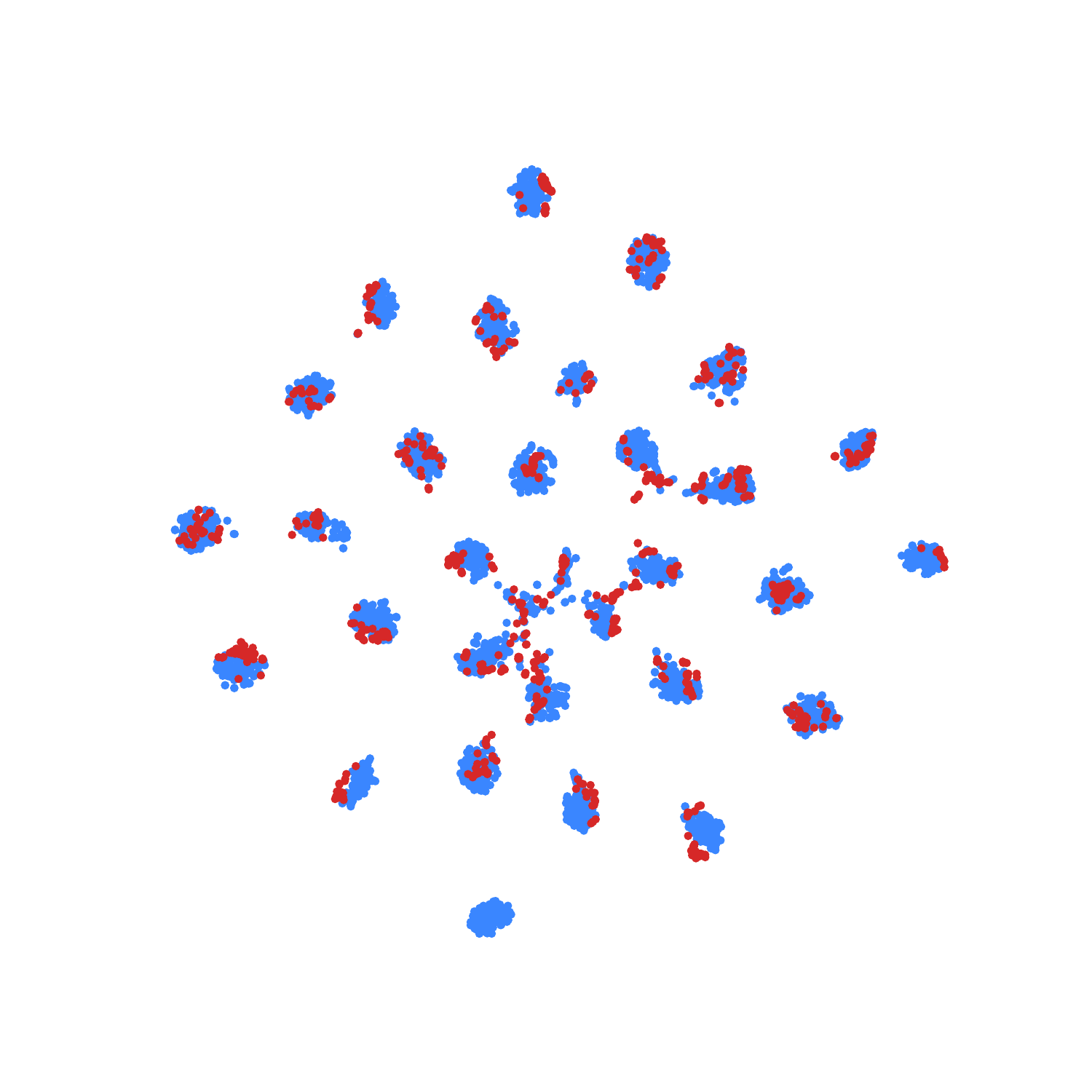}}
    \hfill
    \subfloat[\fontsize{4pt}{4pt}\selectfont GVG-PN]{\includegraphics[width=0.11\textwidth]{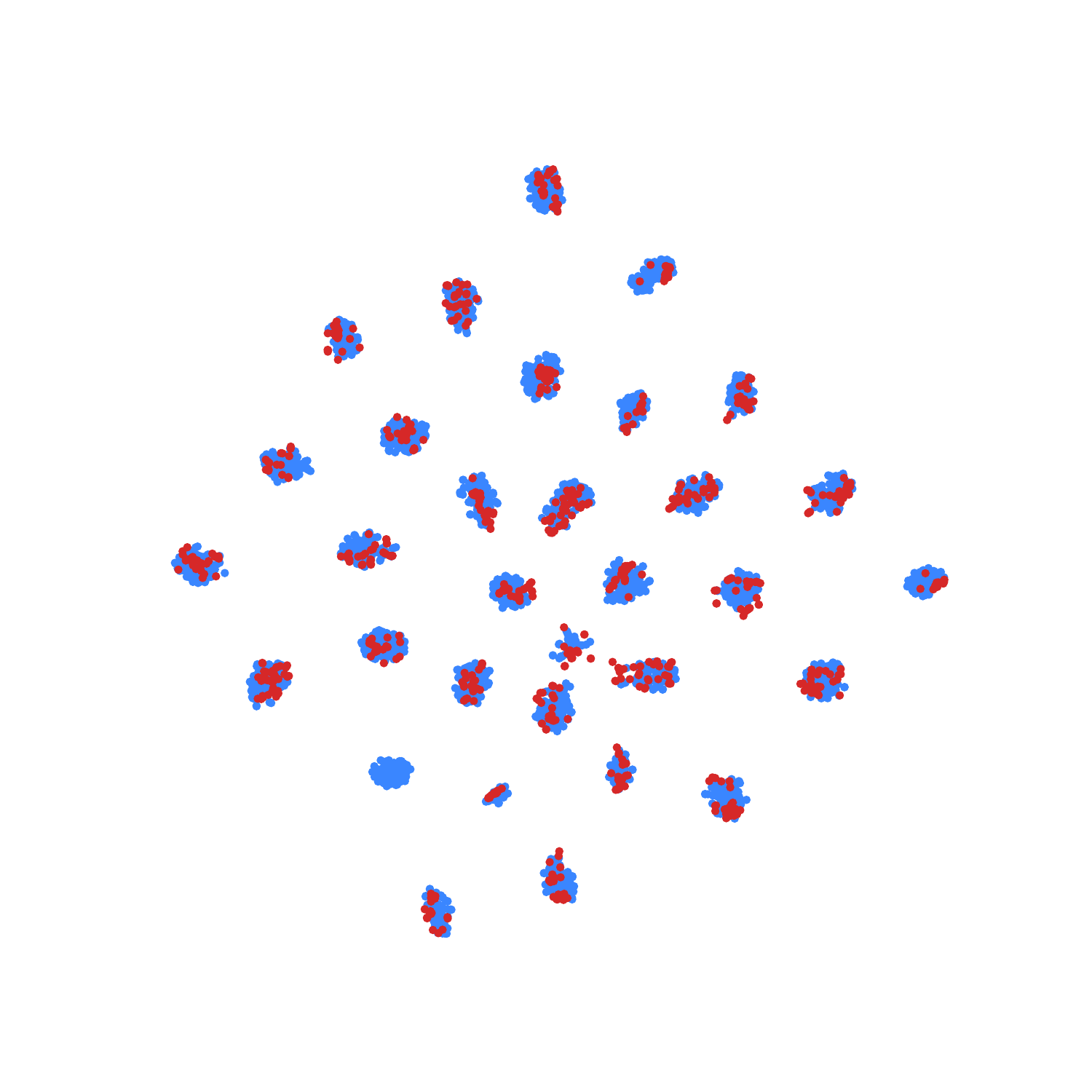}}
    \caption{
Visualizing Embedding Features of Task $A \rightarrow W$ on the Office-31 dataset using the t-SNE Algorithm. \textbf{Top:} Visualizing Clustering of Source and Target Domain Features: (a) Non-adaptation, (b) MSTN, (c) GVG-PN~(w/o $\mathcal{L}_{\text {ProNCE}}$), and (d) GVG-PN. \textbf{Bottom:} Domain Matching Visualization of (e) Non-adaptation, (f) MSTN, (g) GVG-PN~(w/o $\mathcal{L}_{\text {ProNCE}}$), and (h) GVG-PN. Source domain Amazon (\textcolor{blue}{blue}) and target domain Webcam (\textcolor{red}{red}).
}
    \label{tsne}
\end{figure}


\subsubsection{Distribution Discrepancy }
In this section, we investigate the domain adaptation (DA) ability of our model in terms of distributional differences during training. Specifically, we focus on the A $\rightarrow$ W and  W $\rightarrow$ D tasks in the Office-31 dataset and compare the performance of four models: ResNet, MSTN, GVG-PN~(w/o $\mathcal{L}_{\text {ProNCE}}$), and GVG-PN.
To measure the distribution variances, we use the $\mathcal{A}$-distance~\cite{ben2010theory}, which is commonly used in DA. Together with the source risk, the
$\mathcal{A}$-distance constrains the target risk. The $\mathcal{A}$-distance is defined as $d_{\mathcal{A}}=2(1-2 \epsilon)$, where $\epsilon$ represents the error of the binary domain classifier. A larger domain difference corresponds to a larger $\mathcal{A}$-distance.

Figure~\ref{ddcs}(a) presents the distribution differences for the A $\rightarrow$ W and W $\rightarrow$ D tasks in the Office-31 dataset. It is evident that our method effectively reduces the $\mathcal{A}$-distance between domains compared to the other three models.
This demonstrates the effectiveness of our method in aligning the two domains and reducing the domain differences.
Additionally, we observe a significantly smaller $\mathcal{A}$-distance for the W $\rightarrow$ D task compared to the A $\rightarrow$ W task, indicating the high similarity between domains W and D. After adaptation, the classification accuracy reaches  100\%.
Overall, this experiment confirms the superiority of our proposed model in terms of reducing distributional differences and achieving effective domain alignment.
\zh{
\subsubsection{Metrics Analysis of ProNCE}
To investigate the impact of different distance metrics on ProNCE, we introduced the Euclidean distance $\phi (u,v) = \left \| u -v \right \| $ in our experiments as a substitute for Eq.~(\ref{dist}), serving as the metric for ProNCE.
This variant is denoted as GVG-PN\_L2.
In Figure ~\ref{ddcs}(b), we tested the accuracy of GVG-PN\_L2 in the target domain on three tasks within the Office-31 dataset.
It is observed that, although exhibiting some differences compared to GVG-PN, GVG-PN\_L2 demonstrates relatively stable predictions in the target domain.
}
\subsubsection{Convergence}
We evaluated the convergence of different models in the A $\rightarrow$ W task on the Office-31 dataset and plotted the test error curves with respect to the number of iterations in Figure~\ref{ddcs}(c).
It is evident from the plot that the proposed models have lower test errors compared to the other methods. Specifically, our model achieves a test error of 0.043, corresponding to an accuracy of 95.7\%.
Additionally, our model demonstrates faster convergence compared to the other models, indicating its adaptability to the target domain. 
These results highlight the superior convergence properties and effectiveness of our proposed model in achieving accurate adaptation in the A $\rightarrow$ W task.
\zh{
\subsubsection{Pseudo-Labeling Threshold Analysis}
In Figure ~\ref{ddcs}(d), we observe the dynamic changes in the pseudo-label threshold $\delta$ during the training process, which adapts within each mini-batch.
In the early stages, due to the model's limited predictive capability on target domain samples, the threshold is set low.
As training progresses, with the deepening of model knowledge, the threshold gradually increases, eventually stabilizing within a controllable range. 
This variation reflects the ongoing adaptation process of the model to the target domain.
Furthermore, due to the higher accuracy of the model on the A $\rightarrow$ W task compared to the Rw $\rightarrow$ Ar task, the dynamic threshold distribution on the A $\rightarrow$ W task is generally higher than that on the Rw $\rightarrow$ Ar task.
}
\subsubsection{Feature Visualization}
In this section, we utilize t-SNE~\cite{sne} feature visualization
to demonstrate the discriminative and transferable features of the $A \rightarrow W$ task on the Office-31 dataset. This visualization will highlight the advantages of our model compared to other methods. Specifically, we visualize the features of 'Non-adaptation', MSTN, GVG-PN~(w/o $\mathcal{L}_{\text {ProNCE}}$), and our complete model in Figure~\ref{tsne}.
Figure~\ref{tsne}(a)-(d) depict the embedding features, with each category represented by a different color. It is evident from the figures that our complete model exhibits superior discriminative features compared to MSTN and GVG-PN~(w/o $\mathcal{L}_{\text {ProNCE}}$). By leveraging the handling of hard negative pairs through $\mathcal{L}_{\text {ProNCE}}$, GVG-PN demonstrates improved clustering results in the feature space, with only a few hard samples being indistinguishable.

Figure~\ref{tsne}(e)-(h) illustrate the domain matching aspect, showcasing the alignment of features from the two domains. Without adaptation, the features from the source and target domains appear highly disorganized. In MSTN, although class-level alignment is applied, it does not achieve satisfactory alignment between the domains. 
However, the visualizations of GVG-PN~(w/o $\mathcal{L}_{\text {ProNCE}}$) and our complete model indicate that progressively aligning the two domain distributions helps alleviate the domain differences to some extent.
From the results, we can conclude that our model maintains excellent discriminative ability by making the same categories compact and enhancing inter-class separability. Furthermore, it focuses on the hard negative pairs. These results demonstrate that our proposed model outperforms other models in terms of both transferability and discriminability.

\begin{figure*}[htbp]
    \centering
    \subfloat[\fontsize{6pt}{8pt}\selectfont Source-only~(C$\rightarrow$P)]{\includegraphics[width=0.24\textwidth]{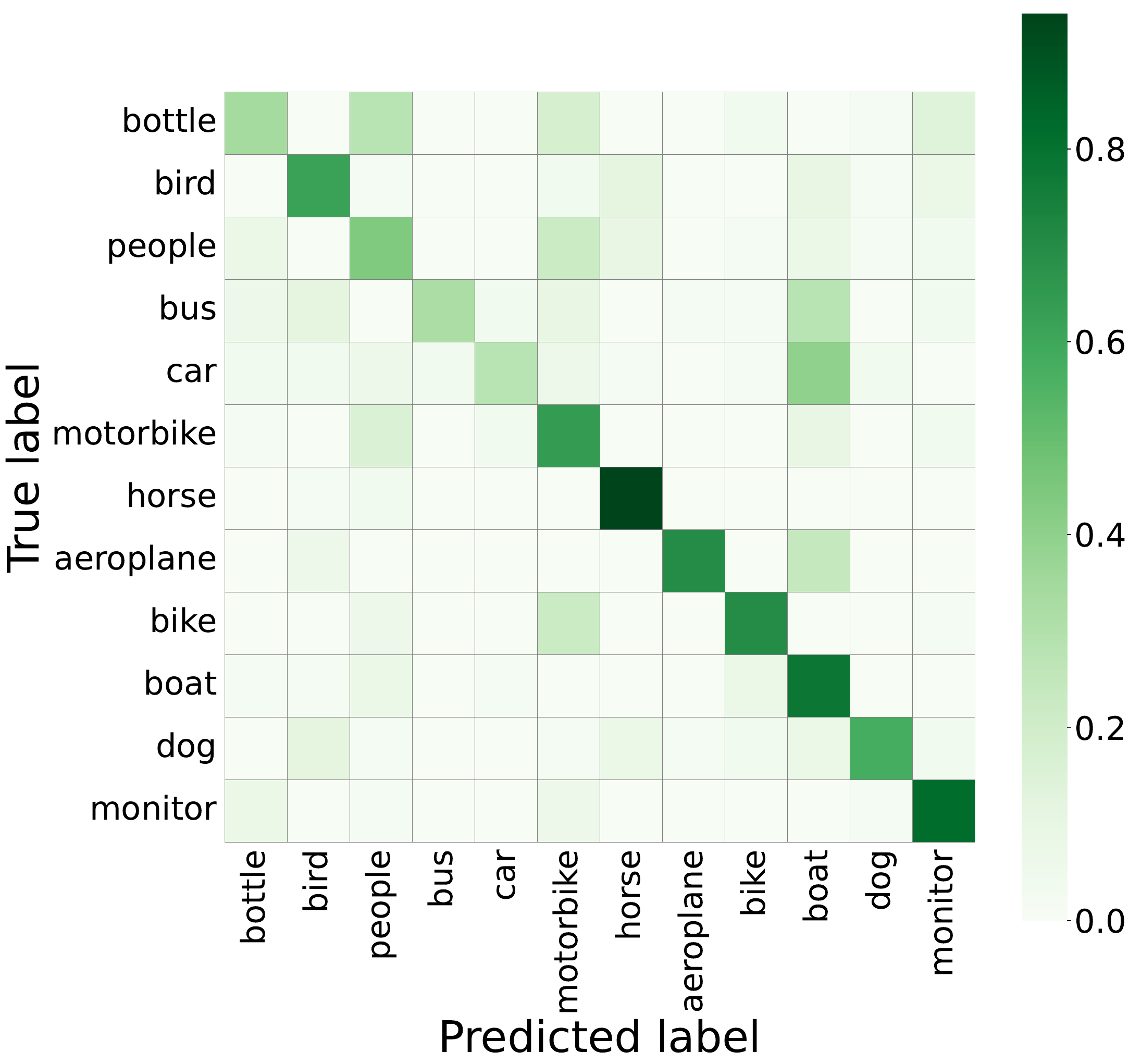}}
    \hfill
    \subfloat[\fontsize{6pt}{8pt}\selectfont GVG-PN~(C$\rightarrow$P)]{\includegraphics[width=0.24\textwidth]{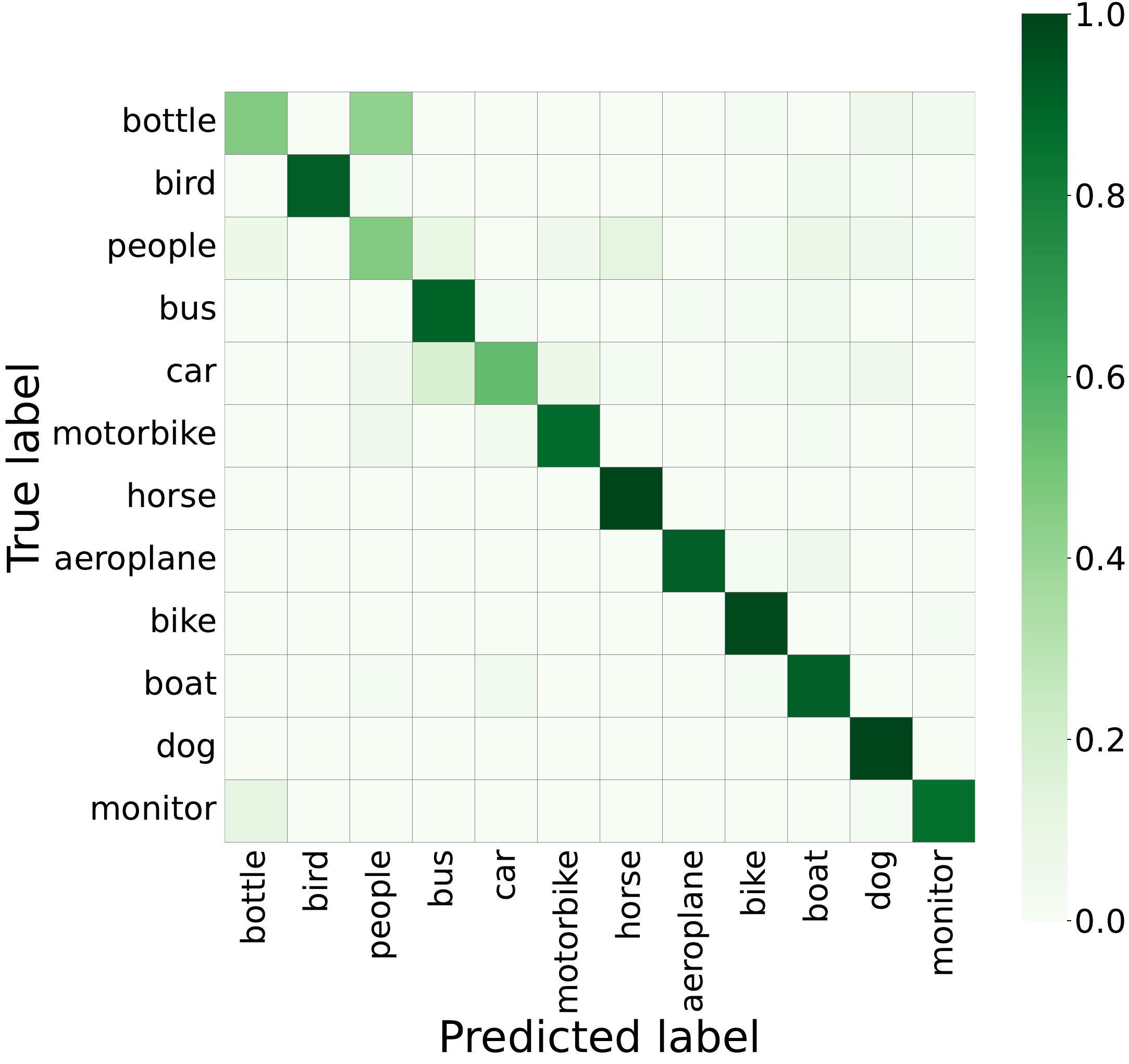}}
    \hfill
    \subfloat[\fontsize{6pt}{8pt}\selectfont Source-only(P$\rightarrow$I)]{\includegraphics[width=0.24\textwidth]{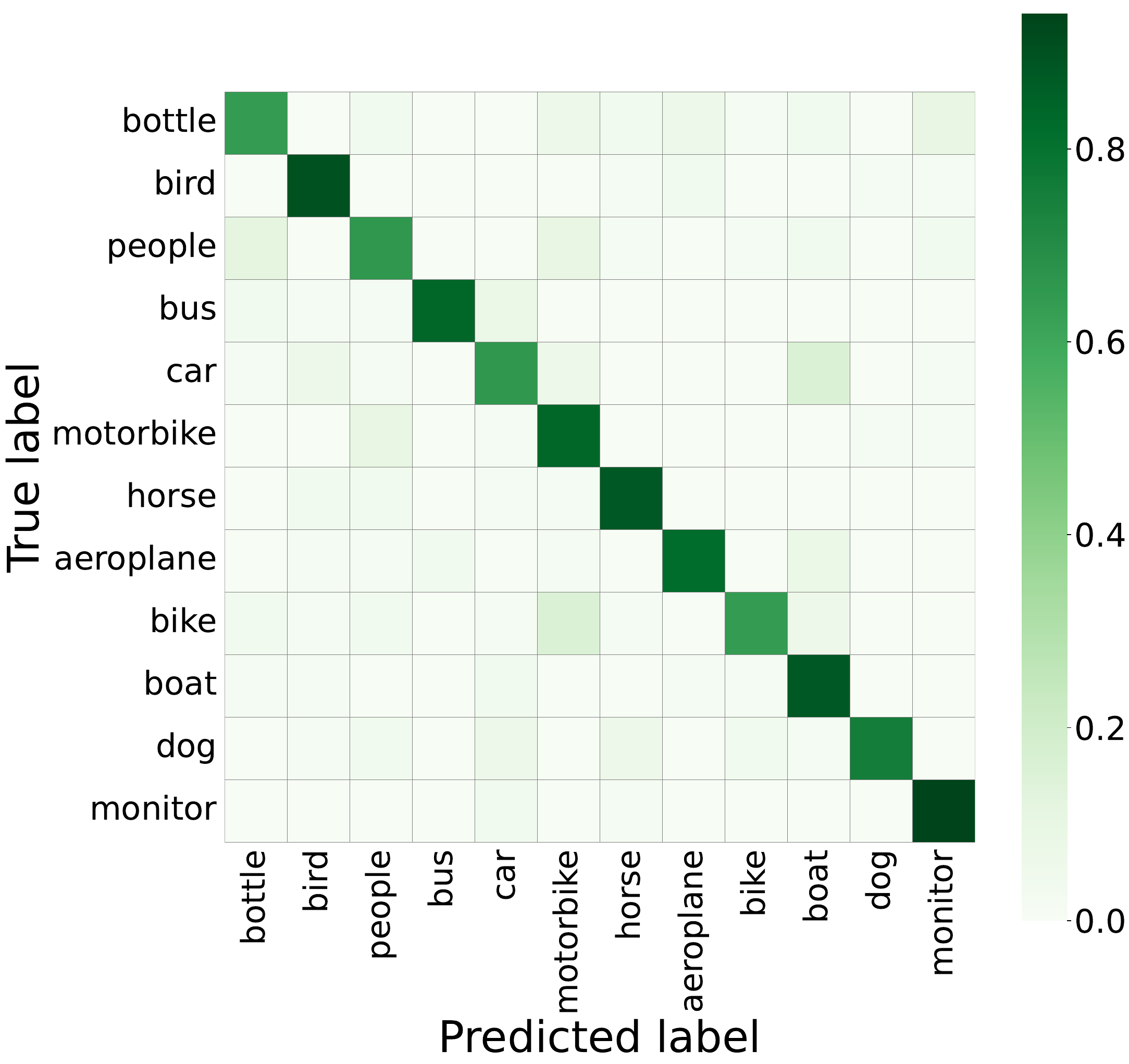}}
    \hfill
    \subfloat[\fontsize{6pt}{8pt}\selectfont GVG-PN~(P$\rightarrow$I)]{\includegraphics[width=0.24\textwidth]{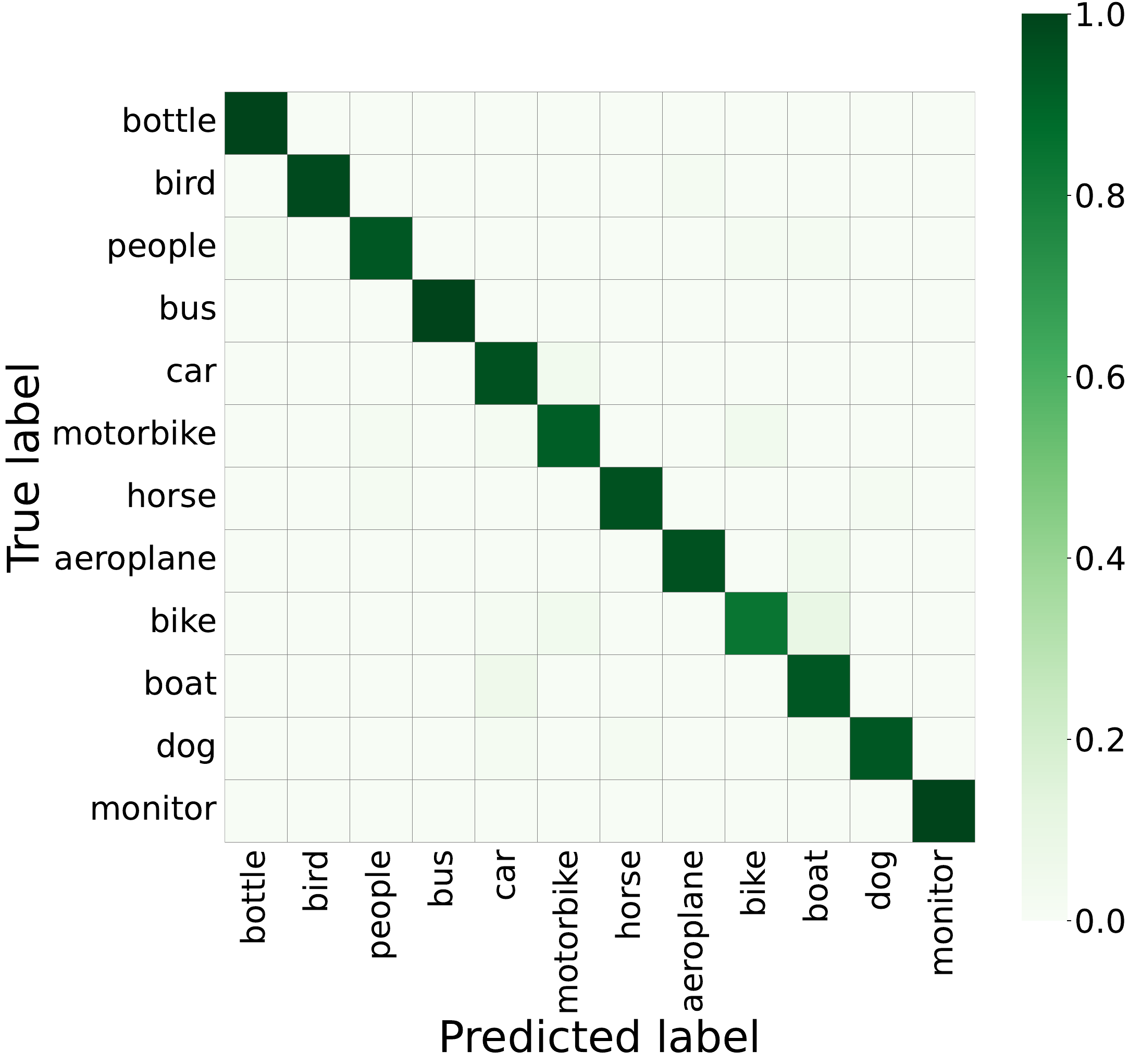}}
    \caption{
The confusion matrix visualization on task C$\rightarrow$P and P$\rightarrow$I in the ImageCLEF-DA dataset, where the horizontal and vertical ordinates denote the true and predicted labels, respectively.
}
    \label{con}
\end{figure*}


\begin{figure*}[htbp]
    \centering
    \subfloat[\fontsize{6pt}{8pt}\selectfont]{\includegraphics[width=0.24\textwidth]{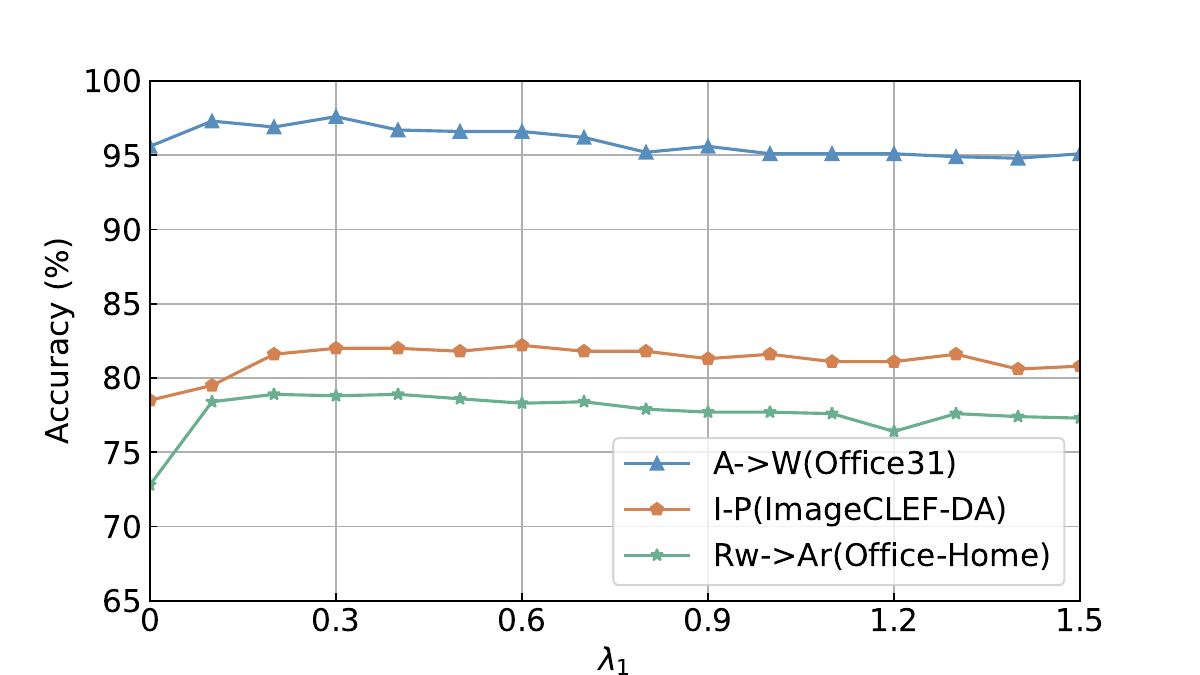}}
    \hfill
    \subfloat[\fontsize{6pt}{8pt}\selectfont]{\includegraphics[width=0.24\textwidth]{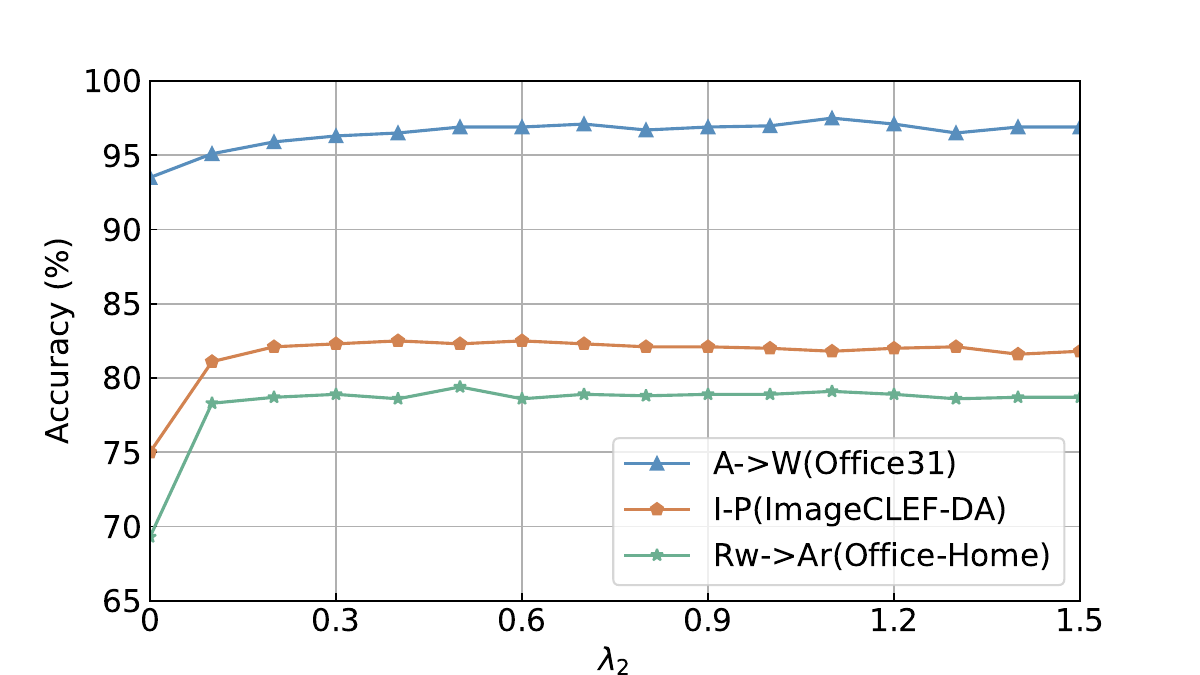}}
    \hfill
    \subfloat[\fontsize{6pt}{8pt}\selectfont ]{\includegraphics[width=0.24\textwidth]{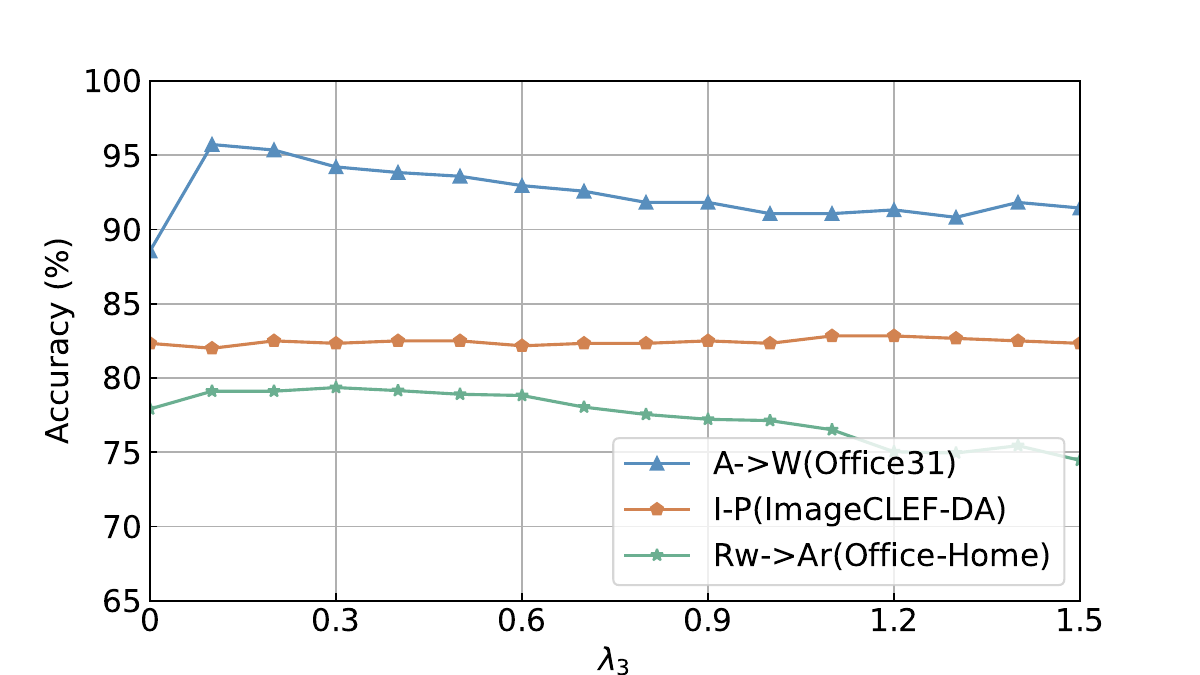}}
    \hfill
    \subfloat[\fontsize{6pt}{8pt}\selectfont ]{\includegraphics[width=0.24\textwidth]{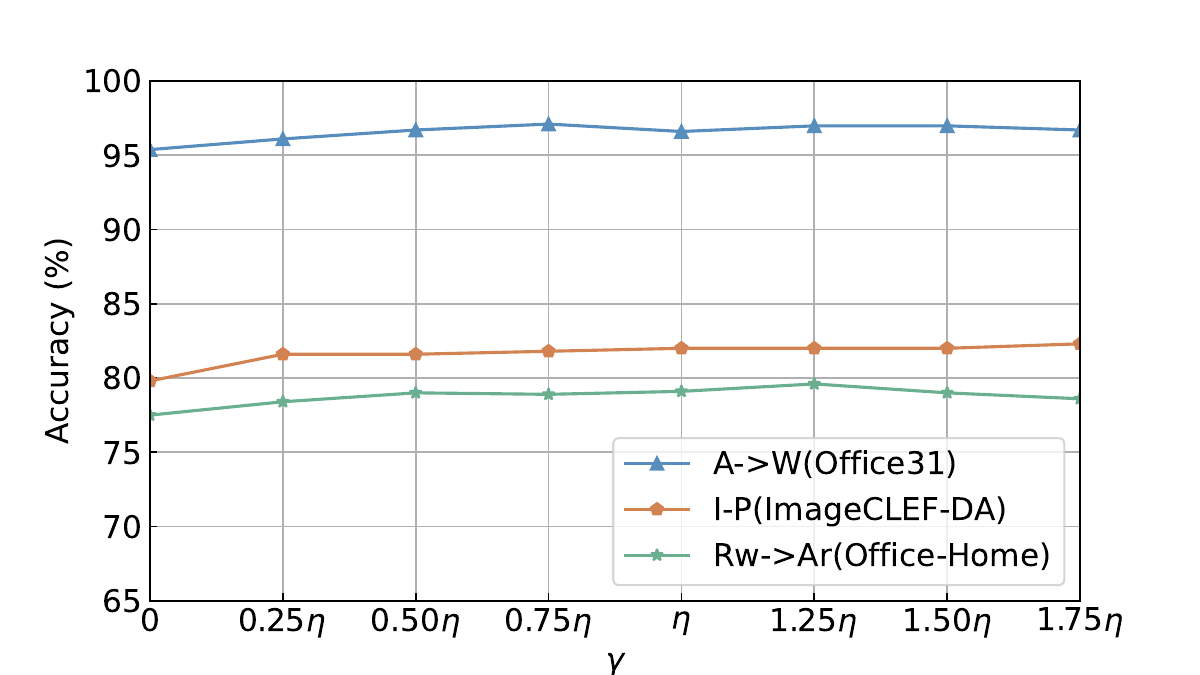}}
    \caption{
We conducted a parameter sensitivity analysis for the GVG-PN model on three transfer tasks: A$\rightarrow$W, I$\rightarrow$P, and Rw$\rightarrow$Ar. (a) classification accuracy versus the variations of $\lambda_1$,(b) classification accuracy versus the variations of $\lambda_2$, 
    (c) classification accuracy versus the variations of $\lambda_3$,
    (d) classification accuracy versus the variations of $\gamma$.
}
    \label{para}
\end{figure*}

\subsubsection{Confusion Matrix Visualization}
We provide the visualization of the confusion matrix in Figure~\ref{con} for the C$\rightarrow$P and P$\rightarrow$I tasks on the
ImageCLEF-DA dataset.
The visualization compares the
 '{Source-only}' approach with our {GVG-PN} method.
 In the 'Source-only' approach, we train the classification model only using the labeled source domain samples and directly apply the model to the target domain. On the other hand, 
our {GVG-PN} method demonstrates the effectiveness of GVG-PN in adapting to the target domain.
As depicted in Figure~\ref{con}(a) and (b), the model trained using the 'Source-only' approach often misclassifies cars as boats, resulting in a significantly lower recognition accuracy for cars compared to the GVG-PN method. 
In the P$\rightarrow$I task, the GVG-PN trained classification model exhibits excellent discriminative power in the target domain.
The visualization of the confusion matrix clearly showcases the improvement in discriminability achieved by our GVG-PN method compared to the 'Source-only' approach.

\subsubsection{Parameter sensitivity analysis}
We conducted a parameter sensitivity analysis to evaluate the robustness of our progressive alignment framework, GVG-PN. The parameters under investigation are $\lambda_1$, $\lambda_2$, $\lambda_3$ and $\gamma$.
We focus on analyzing the balance between generating domain-biased prototypes via GCN and prototype-level contrastive learning.
Figure~\ref{para} illustrates the results of our analysis, demonstrating that our method is relatively insensitive to changes in the parameter values of $\lambda_1$, $\lambda_2$ and $\lambda_3$ within a reasonable range, such as $\lambda_1 \in[0.2,0.9]$, $\lambda_2 \in[0.3,1.5]$ and $\lambda_3 \in[0.1,0.6]$. However, when the values of $\lambda_1$, $\lambda_2$ 
and $\lambda_3$ become too small, there is a significant decrease in model performance. Hence, it is crucial to choose appropriate values within the specified range to ensure the stability of the model performance.

Our proposed domain-biased prototype generation approach can effectively leverage the inter-domain semantic structures and improve the high-level semantic representation of sample features. Based on the domain-biased prototypes, the 
discriminability is further explored through $\mathcal{L}_{\text {ProNCE}}$. 
To analyze the impact of the parameter $\gamma$, we conducted experiments with different values within a specified range. Specifically, we set $\eta=\frac{2}{1+\exp (-\alpha p)}-1$ and tested $\gamma$ values ranging from
$\{0.0, 0.25 \eta, 0.5 \eta, 0.75 \eta, 1\eta, 1.25 \eta, 1.5 \eta, 1.75 \eta\}$.
The results shown in Fig~\ref{para}(d)  indicate that the performance of the model is relatively insensitive to changes in the parameter $\gamma$, implying that the proposed $\mathcal{L}_{\text {ProNCE}}$ is robust and can effectively enhance the model's performance across a range of $\gamma$ values.

\section{CONCLUSION}
In this paper, we proposed a novel progressive domain alignment framework called GVG-PN to address the challenge of negative transfer caused by significant domain differences. In addition to global alignment, our approach focuses on exploring fine-grained semantic structures between domains. 
We generate domain-biased prototypes in intermediate domains based on aggregated sample features to capture domain-specific information. Moreover, we enhance category matching through prototype-level contrastive learning, aiming to improve class-level discriminability.
Through comprehensive experiments and discussions on five benchmark datasets, we clearly validate the effectiveness of our proposed approach. 
\zh{
Our experiments align with existing domain adaptation methods~\cite{li2021transferable,long2015learning}, assuming a balanced distribution of categories, and datasets with only a few dozen or a few hundred categories.
However, this assumption does not hold in specific scenarios where some categories have sparse samples or where certain large-scale datasets have a substantial number of categories.
In our future work, we will investigate the relationship between intermediate domain sample features and domain-bias prototypes, or integrate ideas from existing methods to address these challenges.
}

\bibliographystyle{unsrt}

\end{document}